\documentclass{twinlab}

\usepackage{amssymb}
\usepackage{xspace}
\usepackage{enumitem}
\usepackage{fontawesome5}
\usepackage{wrapstuff}

\usepackage{multirow}
\usepackage{pifont}
\usepackage{tabularx}
\usepackage{makecell}
\usepackage{longtable} %

\graphicspath{{fig_crop/}{media/}}

\usepackage{listings}

\newcommand{\bench}{TwinLab-Bench\xspace}
\newcommand{\prov}[1]{#1}

\renewcommand{\paragraph}[1]{\par\leavevmode{\bfseries #1}\hspace{0.6em}\ignorespaces}


\title{TwinLab-Bench: Evaluating AI Scientists\\ in Agentic Twins of Real Labs}
\shorttitle{TwinLab-Bench}
\shortauthors{Wan et al.}
\renewcommand{\headerbrand}{Preprint}

\renewcommand{\authorlist}{{\sffamily Yang Wan$^{1}$, Linchao Zhu$^{*1}$}}
\renewcommand{\affiliationlist}{{\color{osufg}{\normalsize\sffamily $^{1}$Zhejiang University}\\[3pt]{\small $^{*}$Corresponding author}}}

\newcommand{\projecturl}{https://twinlab-bench.github.io/}
\newcommand{\resourcelink}[3]{{\color{osufg}#1}\hspace{0.35em}{\sffamily\bfseries #2:}~\expandafter\resourcelinkurl\expandafter{#3}\par}
\newcommand{\resourcelinkurl}[1]{\href{\detokenize{#1}}{\texttt{\textcolor{osured}{\detokenize{#1}}}}}
\renewcommand{\infolist}{{\small\color{osufg}%
  \resourcelink{\faGlobe}{Project}{\projecturl}}}

\abstract{AI agents are increasingly expected to act as AI scientists, which requires three increasingly open abilities: carrying out a given computational workflow correctly, choosing the next measurements from a given design space, and designing their own experiments.
Existing benchmarks mainly evaluate the first two abilities, in a terminal or in a closed loop that selects from predefined candidates.
Evaluating the third ability requires reliably executing an experiment that an agent designs, with readings grounded in a real laboratory.
We introduce \textbf{\bench}, which evaluates AI scientists on self-designed experiments in \prov{52} tasks from laboratories in the physical sciences, the life sciences, and engineering.
The benchmark supports more than \prov{100} experimental action types, and the agent's report is scored by comparison with a hidden answer taken from real laboratory records.
Each task has its own TwinLab agent, an agent that serves as the environment and executes the experiments that the tested agent designs, reading the raw materials of the laboratory and calling calculators that are built from its records.
Compared with a language model that sees only the task description, the TwinLab agent almost eliminates readings that deviate from the real records and inconsistent answers to a repeated request, and closed-book probes control for answers recalled from memory.
We find that these tasks remain hard for frontier models: the strongest, Claude Fable 5.2, averages a score of only \prov{45.0}, and we conduct an empirical study of where it fails.
}

\begin{document}

\maketitle

\begin{figure}[h!]
\centering
\includegraphics[width=\textwidth]{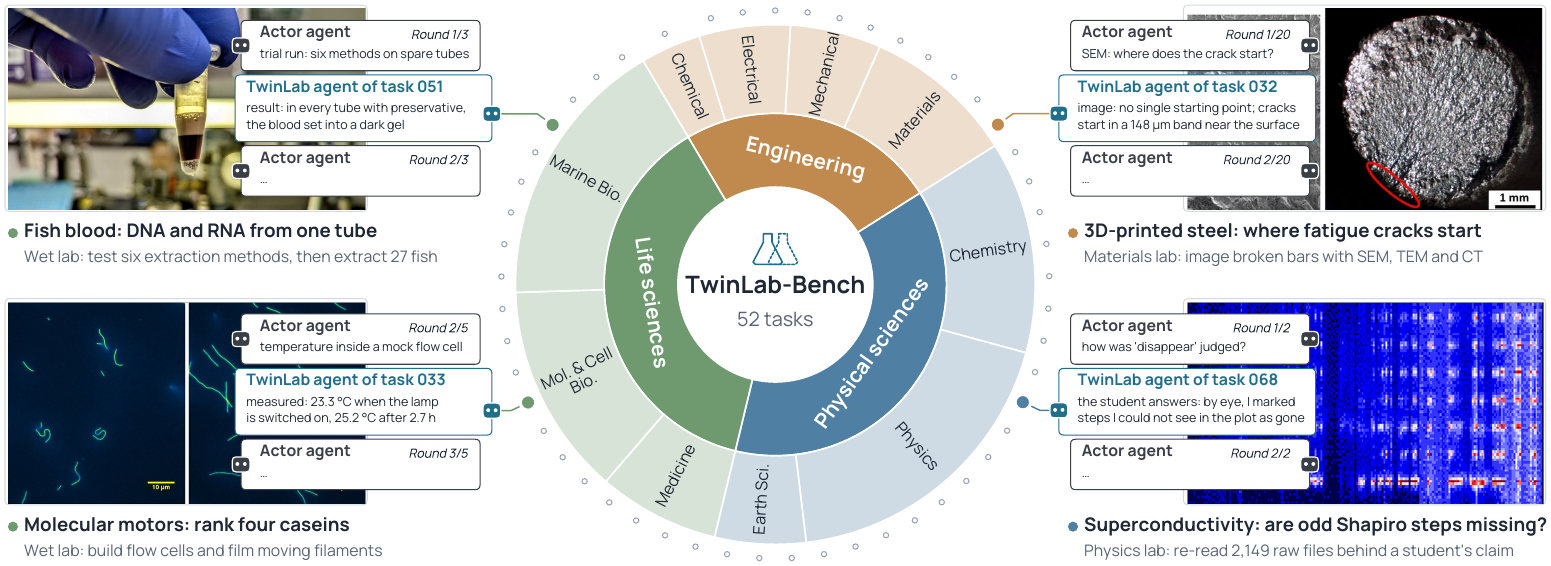}
\caption{TwinLab-Bench covers \prov{52} tasks by division and department, and a separate TwinLab agent drives the environment of each task.}
\label{fig:teaser}
\end{figure}
\clearpage

\section{Introduction}
\label{sec:intro}

\begin{figure*}[t]
\centering
\includegraphics[width=\textwidth]{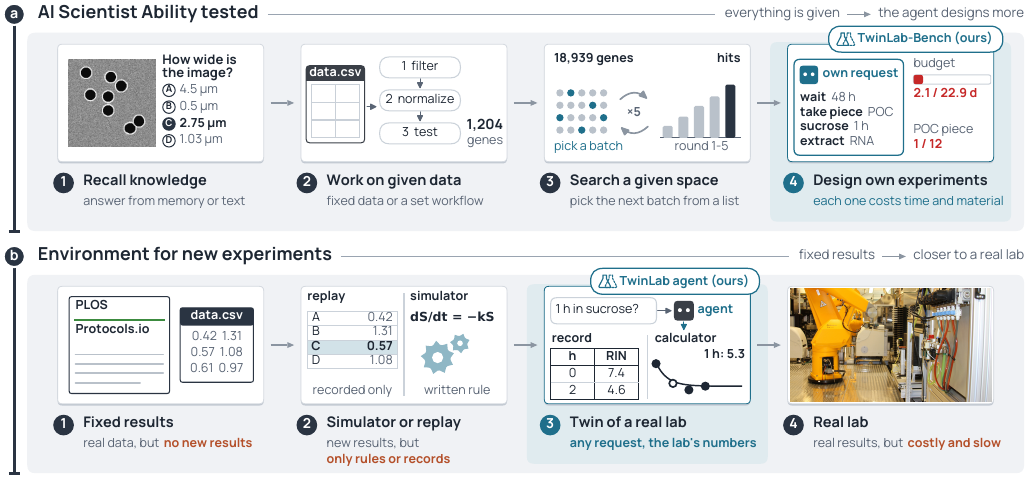}
\caption{\textbf{Two axes for science benchmarks.} (a) From left to right, the agent designs more of the method, and \bench tests ability \ding{175}. (b) From left to right, the environment for new experiments comes closer to a real laboratory, and \bench uses environment \ding{174}.}
\label{fig:axes}
\end{figure*}

\begin{wrapstuff}[type=table,r,width=0.56\textwidth,top=0]
\centering
\small
\setlength{\tabcolsep}{4pt}
\begin{tabular}{@{}lccc@{}}
\toprule
 & \multicolumn{2}{c}{Language model given} & \multirow{3}{*}{\makecell{TwinLab\\agent}} \\
\cmidrule(lr){2-3}
 & \makecell{task\\description} & \makecell{+ raw\\materials} & \\
\midrule
Deviates from the real record & \prov{41\%} & \prov{14\%} & \prov{2\%} \\
Inconsistent when repeated & \prov{31\%} & \prov{8\%} & \prov{4\%} \\
\bottomrule
\end{tabular}
\caption{How often the reply of each environment to a request deviates from the record or changes when the request is repeated; lower is better. Appendix~\ref{app:prelim} gives the details.}
\label{tab:prelim}
\end{wrapstuff}
AI is advancing science in more and more of the research process, and the AI scientists that the field now builds need three increasingly open abilities~\citep{gottweis2026coscientist,ghareeb2026robin}.
The first is carrying out a given computational workflow correctly, such as analyzing a given dataset~\citep{mitchener2025bixbench,chen2024scienceagentbench,tbscience2026,mitchener2025kosmos}.
The second is choosing the next measurements from a given design space~\citep{duan2025scigym,gandhi2025boxinggym,roohani2025biodiscoveryagent,asb2026}.
The third is designing their own experiments, and automated laboratories are starting to run such experiments~\citep{boiko2023coscientist,szymanski2023alab}.
The third ability is the hardest of the three, but few science benchmarks test it.

Existing science benchmarks test whether an agent can recall knowledge~\citep{laurent2024labbench}, work on given data~\citep{mitchener2025bixbench}, or search a given space~\citep{duan2025scigym}, as \cref{fig:axes}(a) shows.
These abilities work on data that the task provides in advance, but the third ability, designing their own experiments, needs an environment that returns real data for new experiments.
\Cref{fig:axes}(b) orders these environments by how close they come to a real laboratory.
Fixed results support no new experiments.
A simulator~\citep{jansen2024discoveryworld,duan2025scigym,gandhi2025boxinggym} or a replay~\citep{roohani2025biodiscoveryagent,asb2026} returns new results, which come only from written rules or from recorded experiments.
The goal of the experiment environment is a real laboratory, which returns real results for any experiment that an agent designs~\citep{bran2024chemcrow,tom2024sdl}.
We move one step from simulators and replays toward this goal with a twin of each real laboratory, and we build \bench on these twins.

\begin{wrapstuff}[type=figure,r,width=0.5\textwidth,top=4]
\centering
\nocite{mandyam2026,chen2017,anthonypetersen2024,poppinga2010notebook,puls2024,snapp2024,zhang2017,whitesamsnotebook,harveylabnotebook}%
\includegraphics[width=\linewidth]{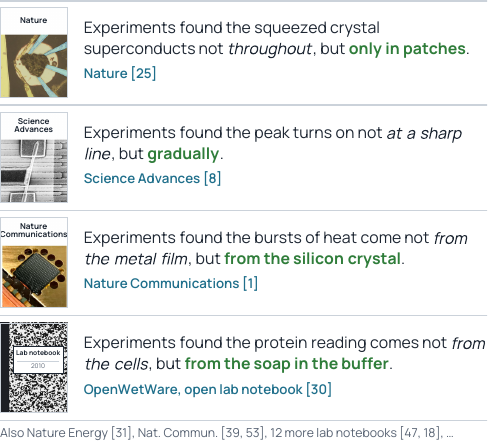}
\caption{\textbf{Task sources.} Four example tasks that come from journal papers or lab notebooks, with the finding that each task targets.}
\label{fig:findings}
\end{wrapstuff}
A twin of a real laboratory has to return real readings for as wide a range of experiments as an agent may design.
Language models already play patients~\citep{schmidgall2024agentclinic} and users~\citep{barres2025tau2} in agent evaluations, but they fall short when they play a laboratory.
Given only the task description, a language model has no basis for the numbers of this laboratory, and given the raw materials, it can still copy wrong numbers.
When we replay the requests that real laboratories made, the language model deviates from the record on \prov{41\%} of these requests with the task description and on \prov{14\%} with the raw materials, as \cref{tab:prelim} shows.

We propose the TwinLab agent to build a reliable twin of each laboratory.
The TwinLab agent is a coding agent that serves as the environment, so it uses tools to read the files of the laboratory and runs programs on them.
Each task has its own TwinLab agent with the materials and the calculators of its laboratory.
For an experiment that the laboratory performed, the TwinLab agent reads the raw materials, often combining several records, and returns the result that the laboratory recorded.
For an experiment that the laboratory did not perform, it runs calculators, which are deterministic programs that we write from the data of this laboratory, for example by fitting a physical model to its measurements.
Each calculator extends the record only to nearby settings, with a basis that we write down when we build the task.

\bench consists of \prov{52} tasks, and each task comes from the records of a real laboratory in the physical sciences, the life sciences, or engineering.
The tested agent, which we call the actor agent, receives the notebooks, raw data, and interim conclusions that the laboratory had at one point of its work, requests the experiments that it designs within a budget, and submits a report that we score by comparing it with what the laboratory later found, as \cref{fig:findings} shows for four tasks.
Our experiments show three properties of \bench.
(1) The score comes from the experiments, because without budget Opus 5.5 and DeepSeek-flash pass none of the tasks.
(2) The actor agent's conclusion follows the true state of the laboratory that it measures.
(3) The twin predicts hidden recorded readings within a median error of \prov{2.47\%} even outside the recorded conditions, and it blocks all \prov{9} attempts to exploit it.

On \bench, the strongest frontier model averages a score of only \prov{45.0} out of 100.
Opus 5.5 still loses part of the score on decisions that it states in full before the run, and its most common decision error is accepting a mechanism without a discriminating test.

Our contributions are threefold.
\begin{itemize}[leftmargin=*,itemsep=1pt,topsep=2pt]
\item \bench, a public benchmark that tests whether an agent can design its own experiments, built on the records of real laboratories and released with its evaluation code.
\item The twin of a real laboratory, made of a TwinLab agent, the raw records of the laboratory, and the calculators that we write from these records.
\item An empirical study that locates where frontier models lose score and which experimental abilities they lack.
\end{itemize}

\begin{figure*}[t]
\centering
\includegraphics{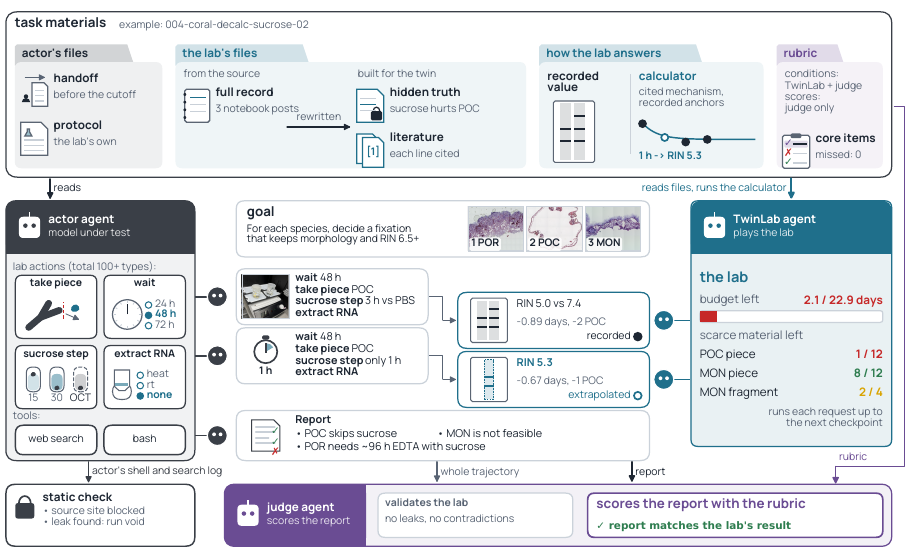}
\caption{\textbf{How a task is run.} The task materials of one example task, and one run with the actor, TwinLab, and judge agents.}
\label{fig:method}
\end{figure*}

\section{Related Work}
\label{sec:related}

\paragraph{Benchmarks without new physical experiments.}
Most science benchmarks give the agent no laboratory for running new experiments.
Question sets test whether an agent recalls scientific knowledge~\citep{rein2024gpqa,phan2026hle,laurent2024labbench,laurent2026labbench2} or how to change a published protocol~\citep{sivakumar2026benchbenchprotocol}, and data-analysis benchmarks hand the agent a dataset and a question~\citep{chen2024scienceagentbench,mitchener2025bixbench}.
AI scientist systems such as Kosmos also automate this kind of analysis at scale~\citep{mitchener2025kosmos}.
Terminal benchmarks~\citep{merrill2026terminalbench,tbscience2026} and benchmarks of automated machine-learning research~\citep{chan2025mlebench,wijk2024rebench,starace2025paperbench} let the agent run new experiments, but these experiments are programs inside a computer.
End-to-end systems already write and review such research~\citep{lu2026aiscientist}, and end-to-end research benchmarks run the agent on the given data and code of a study~\citep{xu2026researchclawbench,zhou2026asibench,yang2026truthinsightbench}.
Benchmarks of these four kinds cannot evaluate how an agent chooses physical experiments that cost time and consume scarce materials.

\paragraph{Readings from simulation, replay, or a language model.}
Interactive environments answer the experimental requests of an agent with readings from a simulation, a replay, or a language model.
Text worlds and equation models~\citep{wang2022scienceworld,jansen2024discoveryworld,duan2025scigym,gandhi2025boxinggym,rowstron2026aria} answer any request, and their readings follow the rules written into the environment.
EmbodiedLab~\citep{gao2026embodiedlab} tests how agents operate instruments in a physics simulator on textbook experiments, while \bench leaves out physical operation and tests experimental decisions on problems from real laboratories.
Replays of published measurements~\citep{roohani2025biodiscoveryagent,asb2026} return real readings, and the agent picks its next experiment among the recorded conditions.
Learned world models predict the outcome of an action~\citep{ha2018worldmodels}, and language models serve as such world models for planning~\citep{hao2023rap} and for text-based games~\citep{wang2024simulators}.
Language models also play patients~\citep{schmidgall2024agentclinic} and users~\citep{yao2025taubench,barres2025tau2} in agent evaluations, and they answer from general knowledge; evaluations of such simulators find that their feedback departs from real environments and users~\citep{liu2026envsimbench,zhou2026sim2real}.
AgentBeats~\citep{agentbeats2026} packages a benchmark as an agent that sets the tasks, the environment, and the scoring for the agent under test.
\bench splits this role between a TwinLab agent that serves as the environment and a judge agent that scores the report.
These environments do not answer the experiments that an agent composes with readings from the records of a real laboratory, so they cannot evaluate how an agent designs its own experiments.

\section{TwinLab-Bench}
\label{sec:method}

\bench consists of \prov{52} tasks, and each task defines a twin of one real laboratory, a budget, and a rubric.
In a run, the tested model, which we call the actor agent, sends experimental requests to the TwinLab agent of the task, an agent that serves as the environment and returns the results, as \cref{fig:method} shows.
The actor agent finally submits a report, and a judge agent scores this report by comparing it with the hidden truth.
All TwinLab agents run the same program, and the materials and the calculators of each task make it the twin of its laboratory, with the engineering details in Appendix~\ref{app:mech}.

\subsection{Tasks}
\label{sec:task}

\paragraph{Task sources.}
We select the tasks from about \prov{375} candidate sources, and each task is built on a real research record, such as an open laboratory notebook, raw instrument exports, a paper with its supplementary material, or a public dataset, as \cref{fig:findings} shows.
Most tasks are adapted from the records of one laboratory.
In some of them the laboratory itself met the problem and later solved it, and in the others the problem lies in the real data of the laboratory.
Their answers come from a later notebook entry, a publication, a correction, or a reanalysis of the raw data.
The few remaining tasks are grounded syntheses, as Appendix~\ref{app:tasks} describes.

\paragraph{Task construction.}
We choose a cutoff point in the record, which is a date or a clearly defined state of the laboratory, and split the record into two parts.
The data, protocols, and interim conclusions that the laboratory had before the cutoff point are handed to the actor agent.
What the source records show after the cutoff point is given only to the TwinLab agent and to the judge agent, and we call this part the hidden truth.
An expert route taken from the record averages a score of \prov{97.8} out of 100 when we replay it in the twin of each task.
Every task also passes a coverage review, which confirms that the standard experiments of the field can be simulated, and Appendix~\ref{app:tasks} lists all tasks and their construction.

\begin{figure*}[t]
\centering
\includegraphics[width=\textwidth]{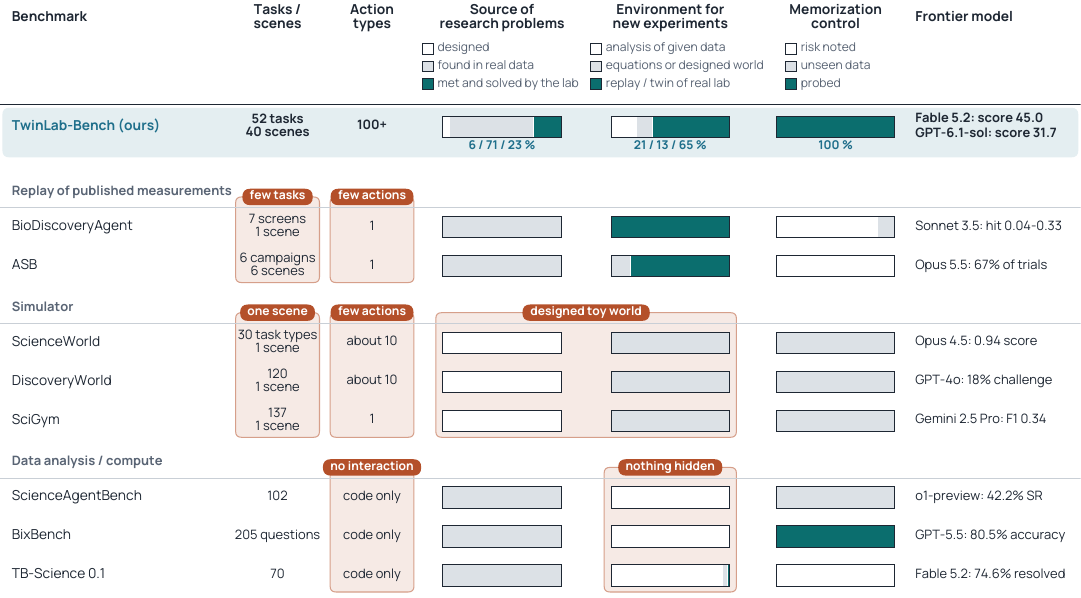}
\caption{\textbf{Scorecard of science benchmarks.} Each row is a benchmark, and each bar splits its tasks by the levels in the key of its column. Orange boxes mark the main weakness of each group of benchmarks.}
\label{fig:scorecard}
\end{figure*}

\paragraph{Task forms.}
The tasks take three forms.
Most tasks are claim reviews, which ask the actor agent to decide which conclusions of the laboratory hold and how to test them.
Troubleshooting tasks ask for the cause of a clear failure and a working procedure, and research-campaign tasks ask for an executable procedure toward a research goal.
Some tasks offer no experiments, and the actor agent analyzes only the raw records that the laboratory already has.

\begin{wrapstuff}[type=table,r,width=0.52\textwidth,top=2]
\centering
\small
\setlength{\tabcolsep}{3pt}
\begin{tabular}{@{}lccc@{}}
\toprule
 & Actor & TwinLab & Judge \\
\midrule
Handoff before the cutoff & \checkmark & \checkmark & \checkmark \\
Full record and raw files & & \checkmark & \checkmark \\
Hidden truth & & \checkmark & \checkmark \\
Rubric full-credit conditions & & \checkmark & \checkmark \\
Maximum score of each rubric item & & & \checkmark \\
Actor's shell and search log & \checkmark & & \checkmark \\
How each reading was produced & & \checkmark & \checkmark \\
\bottomrule
\end{tabular}
\caption{\textbf{Who sees what.} Materials that each agent of a run sees.}
\label{tab:visibility}
\end{wrapstuff}
\paragraph{Actor agent.}
The actor agent proposes its own hypotheses and composes its own experiments.
It describes each experiment in natural language and decides which actions to run, on which materials, with which settings, in which order, and how many times.
Across the benchmark the laboratories support more than \prov{100} action types.
Each laboratory lists its procedures and the parameter options of its equipment and reagents, and the actor agent combines them into experiments and can also request values outside the list.
These lists contain no experiment and no step of the solution, as Appendix~\ref{app:mech} details.
In its sandbox, the actor agent can analyze the handoff materials freely with a shell and Python, but new readings come only from experimental requests to the TwinLab agent, and every request costs budget.
The actor agent can search the web with the source record blocked, and it finally submits a report with its conclusions and recommended procedure.

\paragraph{Difficulty control.}
We control the difficulty of the tasks with five properties that make them depend on experimental judgment.
(1) The budget of each task, counted in units such as laboratory days or wells, is one to three times the amount that the expert route uses for most tasks, so it covers the expert route and a few plausible hypotheses but not all options.
(2) Scarce materials named in the record cannot be recovered once used.
(3) An artifact has a plausible benign explanation.
(4) A measurement that looks decisive has another possible cause, for instance when all samples measured on one date also come from one batch.
(5) In some tasks the correct answer is that the question cannot be decided within the budget.
Overall, for tasks adapted from real records, these five situations are taken from the record, and we add no problem to the record.

\subsection{The TwinLab Agent}
\label{sec:twin}

The environment of each task is a twin of one specific laboratory, made of a TwinLab agent and the task materials of this laboratory.
The task materials contain the full record of the laboratory with its raw files, the calculators that we write from this record, and the hidden truth.
The TwinLab agent handles each experimental request in three steps.
First, it looks for a basis in the task materials, often combining several records.
Second, it takes the recorded value of a condition that the record contains on comparable material or runs calculator commands in a shell.
Third, it writes these numbers as an observation in the instrument's own format and returns a matching real image when one exists.
The environment code writes a session record after every request, so the TwinLab agent keeps the state that earlier requests left, such as a sample already used up.
Its tools only read files, search text, and run the calculators.

A calculator is a deterministic program written for one task from the laboratory's own data, for example by resampling the recorded measurements, by fitting a physical model to the record, or by filling a real instrument's export template.
The TwinLab agent follows two further rules.
\begin{itemize}[leftmargin=*,itemsep=1pt,topsep=2pt]
\item \textbf{Noise.} Independent samples receive noise at the measured repeatability.
\item \textbf{Simulation limits.} A quantity that neither the record nor the literature constrains is marked as a simulation limit, costs nothing, and is reported as not being an experimental result.
\end{itemize}

\begin{figure*}[t]
\centering
\includegraphics[width=\textwidth]{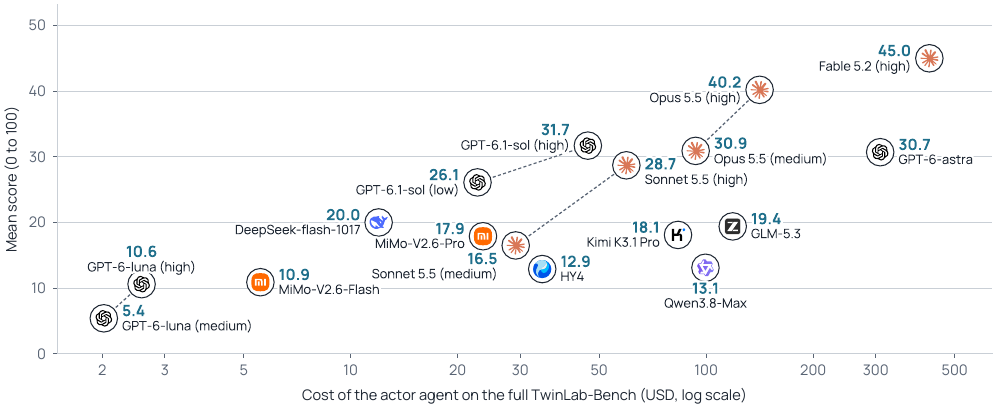}
\caption{\textbf{Mean score versus the cost of the actor agent}; models run at high effort by default.}
\label{fig:main}
\end{figure*}

\paragraph{Local extension.}
The calculators extend the record only locally.
An extrapolated value carries a basis that we establish through a literature search when the task is built, typically an interpolation between two recorded settings or a dependence documented in the literature, and this basis is written into the task item by item.
\Cref{sec:fidelity} tests how much the results depend on these extensions.

\paragraph{Hidden truth.}
The TwinLab agent needs the hidden truth because the readings have to agree with the true state of the laboratory, such as a cause that only appears later in the record.
It answers only what an experiment returns, and Appendix~\ref{app:mech} lists the rules that the environment code enforces.
\Cref{tab:visibility} lists what each role can see.

\subsection{Scoring}
\label{sec:scoring}

\paragraph{Rubric.}
The rubric of each task consists of items with conditions for full and partial credit~\citep{starace2025paperbench,arora2025healthbench}.
Most items belong to two components. Conclusion items check whether the report states the correct answer, and verification items check whether the report reaches this answer from evidence that the actor agent obtained in this run, so a key finding of a task is often split into one item of each kind.

\paragraph{Score and judge.}
The score of each task is rescaled to 0 to 100, and each task marks a few core items whose failure makes the report wrong; when any core item misses full credit, the run scores 0.
We report the mean score over all tasks.
The judge agent is a vision-language model that serves as the judge~\citep{zheng2023judging}, reads the full task, the interaction log, and the report, and cites the report text or the request number for every item.
Every core item with full credit is checked twice more by independent calls, and the item scores 0 when either check fails.
We freeze the rubric before looking at the final evaluation runs.

\paragraph{Invalid runs.}
The judge agent also audits the environment, and a run is invalid when a reading leaks the hidden truth, contradicts earlier observations or physics in a way that matters for scoring, or refuses a relevant experiment that a real laboratory can run.
A deterministic script replays every calculator command to check the state, the billing, and the numbers.
A static check scans the content that reaches the actor agent through its tools, and a run that receives a blocked string of the source is invalid, as Appendix~\ref{app:mech} details.

\section{Experiments}
\label{sec:exp}

\subsection{Comparison with Existing Benchmarks}
\label{sec:compare}
The scorecard of \cref{fig:scorecard} compares \bench with \prov{8} existing benchmarks item by item on scale and diversity, realism, and memorization control, and \bench is the only benchmark in the comparison that combines a wide range of experiments with real problems and real readings, as Appendix~\ref{app:scorecard} discusses.

\subsection{Main Results}
\label{sec:main}

\paragraph{Setup.}
We evaluate \prov{13} models as actor agents, running the GPT models in Codex and the other models in Claude Code.
The TwinLab agent and the judge agent run on DeepSeek-flash for every actor agent.
We report cost together with score following \citet{kapoor2024agents}; \cref{fig:main} counts the cost of the actor agent, and the TwinLab agent and the judge agent add about \prov{55} USD for all tasks, as \cref{tab:env-cost} shows.

\paragraph{Results.}
As \cref{fig:main} shows, Claude Fable 5.2 scores \prov{45.0} on average and Claude Opus 5.5 scores \prov{40.2}, and the other actor agents score between \prov{5.4} and \prov{31.7}.
\Cref{tab:main} lists the scores by division and task form.
For \prov{127} of the \prov{136} pairs of actor agents, with each reasoning effort of a model counted as one actor agent, the actor agent with the higher mean score also scores higher on more tasks than it scores lower, and a paired comparison over tasks separates \prov{66} of the \prov{75} pairs whose mean scores differ by at least 10, as Appendix~\ref{app:ds-repeats} shows.
We find that (1) the score tends to rise with the logarithm of the cost, and both a larger model and a higher reasoning effort raise the score, and (2) the differences between actor agents hold across many tasks.

\paragraph{Task scores separate correct and wrong final reports.}
With a fixed rule, we select \prov{21} tasks whose final reports include both correct and wrong ones, which we divide by the hidden truth without looking at their scores.
In \prov{19} of these \prov{21} tasks, every correct report has a higher total score than every wrong report, by \prov{0.46} of the maximum score on average, as Appendix~\ref{app:separation} shows.
The judge agent is stable, repeating a judgment has little effect, and two judge agents agree on the pass verdict in \prov{19} of \prov{20} runs, as Appendix~\ref{app:judge-stab} and \ref{app:swap} show.

\subsection{Fidelity of the Twin}
\label{sec:fidelity}

We test along a chain of evidence how far the results that the twin returns can be trusted.

\paragraph{Most experiment results come from real laboratory data.}
We trace where the experiment results of the actor agents come from and whether the record contains the decisive experiments.
Of these experiment results, \prov{76\%} come from the record or from calculators calibrated on the same laboratory's data, as Appendix~\ref{app:provenance} shows, and the record contains the conditions of \prov{78\%} of the decisive experiments on the expert routes, as Appendix~\ref{app:extrap-audit} shows.
Thus the conclusions of most tasks depend on recorded data, and the calculators mostly fill in details around them.

\paragraph{Sensitivity to extrapolation.}
We replace the extrapolation outside the record with another equally plausible choice and rerun the actor agents on six tasks, where the replacement changes \prov{12\%} to \prov{44\%} of the calculator outputs, as Appendix~\ref{app:extrap-sens} details.
We find that most conclusions do not depend on which plausible extrapolation the calculators use, so a modeling choice that we make when we build a task changes few evaluation outcomes.

\paragraph{Held-out conditions.}
We hide recorded conditions of three tasks and rebuild their calculators without seeing them.
The median error of the hidden readings is \prov{0.74\%} for interpolated conditions and \prov{2.47\%} for extrapolated ones, and a hidden discrete condition is predicted within \prov{0.6\%} on average, as Appendix~\ref{app:holdout} shows.
We find that most readings that the calculators compute near the record fall close to the real readings, so the twin returns realistic readings for new experiments near the record.

\paragraph{The twin supports routes beyond the expert route.}
In successful runs, more than a third of the requests are new experiments that the expert route does not contain, such as new combinations of settings and additional checks, as Appendix~\ref{app:routes} shows.
The TwinLab agent answers these requests, so actor agents can solve a task along their own routes.

\paragraph{Robustness of the twin against hacking.}
When we tell Opus 5.5 on four tasks that the laboratory is simulated and that it may exploit any weakness, all \prov{9} exploit attempts are blocked, and when we ask it to find flaws in the simulation, none of the \prov{13} true flaws that it finds changes the scoring, as Appendix~\ref{app:redteam} details.
An audit of the \prov{7808} replies of the TwinLab agent in \prov{831} evaluation runs finds one reply that states part of the hidden truth and that the judge agent did not flag, and that run scores 0, as Appendix~\ref{app:provenance} shows.
These results show that the score is hard to raise through the weaknesses of the twin and that the flaws that exist are minor.

\begin{table*}[t]
\centering
\begin{minipage}[t]{0.46\textwidth}\vspace{0pt}
\centering
\centering
\small
\begin{tabular}{@{}lc@{}}
\toprule
Actor agent's conclusion in the new state & States \\
\midrule
Follows the new state & \prov{8} \\
Follows it in direction only & \prov{1} \\
Consistent, not verified & \prov{1} \\
Never touched the changed part & \prov{1} \\
\bottomrule
\end{tabular}
\caption{\textbf{Actor agents' conclusions when the true state changes.} We keep the handoff word for word and change only the true state of the laboratory.}
\label{tab:altstate}
\end{minipage}\hfill
\begin{minipage}[t]{0.5\textwidth}\vspace{0pt}
\centering
\centering
\includegraphics[width=\linewidth]{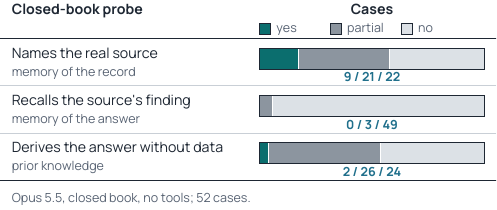}
\caption{\textbf{Closed-book memory probes} of Opus 5.5 on all tasks. Bars give the number of tasks with a yes, partial, or no answer.}
\label{tab:memory}
\end{minipage}
\end{table*}

\subsection{Memorization Control}
\label{sec:memory}

The tasks of \bench come from public records, so a model may have seen a source during training~\citep{sainz2023contamination,golchin2024timetravel} or may derive part of an answer from prior knowledge, and we test in three ways whether the scores come from the agent's own experiments.

\paragraph{Changing the true state of the laboratory.}
To test whether the answer is determined by the readings that the actor agent obtains in its own experiments, we change the true state of the laboratory and keep the handoff word for word.
We construct the new state from a documented case of the same kind of problem by changing only the calculator and the hidden truth, and a model that remembers the source answers according to the source~\citep{wu2024reasoning}.
In \prov{8} of the \prov{10} states that allow a verdict, the conclusion follows the new state, and no run mentions the source study, as \cref{tab:altstate} and Appendix~\ref{app:altstate} show.

\paragraph{Closed-book memory probes.}
In a closed-book session without tools, Opus 5.5 recalls the conclusion of the source for none of the tasks and derives the answer without data for only \prov{2} tasks, and Fable 5.2 recalls the conclusion for one task, as \cref{tab:memory} and Appendix~\ref{app:memo} show.
The tasks whose source Fable 5.2 names do not score higher in its evaluation runs.
Prior knowledge gives at most part of an answer on many tasks.

\paragraph{Pass rate without experiments.}
With a budget of zero, Opus 5.5 and DeepSeek-flash pass none of the tasks, as Appendix~\ref{app:zero} shows.
These results show that prior knowledge and the handoff do not earn the core items, and the agent has to obtain the evidence through its own experiments.

\begin{table*}[t]
\centering
\begin{minipage}[t]{0.48\textwidth}\vspace{0pt}
\centering
\centering
\small
\begin{tabular}{@{}lcc@{}}
\toprule
Before the run, the plan & \makecell{Score expected\\from the plan} & \makecell{Score in\\the run} \\
\midrule
States the decision in full & full & \prov{87\%} \\
States only the principle & partial & \prov{66\%} \\
Does not mention it & none & \prov{62\%} \\
\bottomrule
\end{tabular}
\caption{\textbf{Knowing-doing gap in experimental decisions.} Rows group the key decisions by how the plan states them. Opus 5.5 writes the plan before each run, and each percentage is the score that the run obtains on these decisions out of their maximum.}
\label{tab:decision}
\end{minipage}\hfill
\begin{minipage}[t]{0.48\textwidth}\vspace{0pt}
\centering
\centering
\small
\setlength{\tabcolsep}{4pt}
\begin{tabular}{@{}lc@{}}
\toprule
Failure type & Runs \\
\midrule
\multicolumn{2}{@{}l}{\textit{Experimental decision}} \\
\hspace{0.6em}Accepts a mechanism without testing it & \prov{7} \\
\hspace{0.6em}Chooses a procedure on weak early evidence & \prov{4} \\
\hspace{0.6em}Uses up a scarce material before testing & \prov{1} \\
\textit{Reading the results or reporting} & \prov{4} \\
\midrule
Runs that score 0 & \prov{16} \\
\bottomrule
\end{tabular}
\caption{\textbf{Where runs fail.} Each run of Opus 5.5 that scores 0 on the tasks that need experiments is counted once, under the error that caused it.}
\label{tab:failtype}
\end{minipage}
\end{table*}

\subsection{Empirical Analysis of Model Failures}
\label{sec:failure}

On the tasks that need experiments, the decisions that a model states before the run match only in part the decisions that it makes during the experiments, similar to the knowing-doing gap that language-model agents show in bandit tasks~\citep{schmied2025greedy}.

\paragraph{Knowing-doing gap in experimental decisions.}
Before each run, the model reads the handoff files and writes a plan for the whole task, and we check whether the plan states each key decision of the task in full, as \cref{tab:decision} shows.
Decisions that the plan states in full score \prov{87\%} of their maximum in the run, but decisions that it does not mention still score \prov{62\%}.
Of the \prov{11} decisions that the plan states in full and the run still loses, the run carries out \prov{9} as planned and loses the score in reading its own results or in the report.
A static benchmark that grades the plan therefore misjudges many of these decisions, and a closed-book question gives the same order of scores, as Appendix~\ref{app:decision} shows.

\paragraph{Failure types of the actor agents.}
We classify every valid run of Opus 5.5 that scores 0 by the error that caused the failure, as \cref{tab:failtype} and Appendix~\ref{app:failure} report.
Of the \prov{16} failures of Opus 5.5, \prov{12} come from an experimental decision, most often accepting a mechanism without a discriminating test.
For example, when an oyster sugar assay reads about \prov{30\%} low, Opus 5.5 rules out the pipettes after a water check and never tests the \prov{1}~$\mu$L transfer, which delivers only about \prov{70\%} of the sample.

\paragraph{More budget rarely helps.}
Of the \prov{21} failed runs of Opus 5.5, only \prov{8} spend at least \prov{90\%} of the budget.
Doubling their budget turns only \prov{2} of them into a pass, as Appendix~\ref{app:budget} shows.

\section{Conclusion}
\label{sec:conclusion}
\bench evaluates whether an AI scientist can design its own experiments on twins of real laboratories, and most experiment results in these twins come from real laboratory data.
The strongest frontier model averages a score below 50 out of 100.
Opus 5.5 loses score even on decisions that it states in full before the run, and its most common experimental error is accepting a mechanism without a discriminating test.
We release the tasks and the evaluation code, and the web search of the actor agent blocks the names of the benchmark and its tasks.

\section*{LLM Usage Statement}
Language models are part of the method of this work. The actor agents are the tested models of \cref{sec:main}, and the TwinLab agent, the judge agent, and the analysis calls of \cref{sec:exp} run on the models named there.
We build the tasks with a coding agent (Claude Code with \prov{Claude Opus 5.5}) under our direction.
We also use Claude Code with Claude Opus 5.5 to draft and revise the text, to write the analysis and plotting scripts, and to check the numbers in the text against the results.
The authors take full responsibility for all content of the paper.

\bibliographystyle{twinlab}
\bibliography{main}

\clearpage
\beginappendix

\section{Full Results, Divisions, and Repeated Runs}
\label{app:ds-repeats}

\Cref{tab:main} gives the numbers behind \cref{fig:main}, with one row per actor agent and reasoning effort.
Besides the mean over all tasks, it lists the mean by academic division (\prov{20} tasks in the physical sciences, \prov{20} in the life sciences, and \prov{12} in engineering), the mean by task form, and the cost of the actor agent on all tasks; \cref{tab:env-cost} gives the costs of the TwinLab agent and the judge agent.
The cost of the models other than Claude is computed from their token counts at the list prices of their providers; for runs whose actor agent session ended without a usage record, the actor agent cost is taken as the mean of the recorded runs of the same model.
The prices of GPT-6.1-sol and GPT-6-astra are higher for inputs of \prov{272k} tokens or more, and we apply the tier for inputs below \prov{272k} tokens to every request.
In USD per million tokens of uncached input, cached input, cache write, and output, this tier is \prov{2.00}, \prov{0.10}, \prov{2.50}, and \prov{10.00} for GPT-6.1-sol and \prov{10.00}, \prov{1.00}, \prov{12.50}, and \prov{50.00} for GPT-6-astra.
Thirteen runs on the radiocarbon task were judged under the rubric in force when they ran, and the later change of this rubric only widens the full-credit condition of one item.

\begin{table*}[t]
\centering
\includegraphics[width=\textwidth]{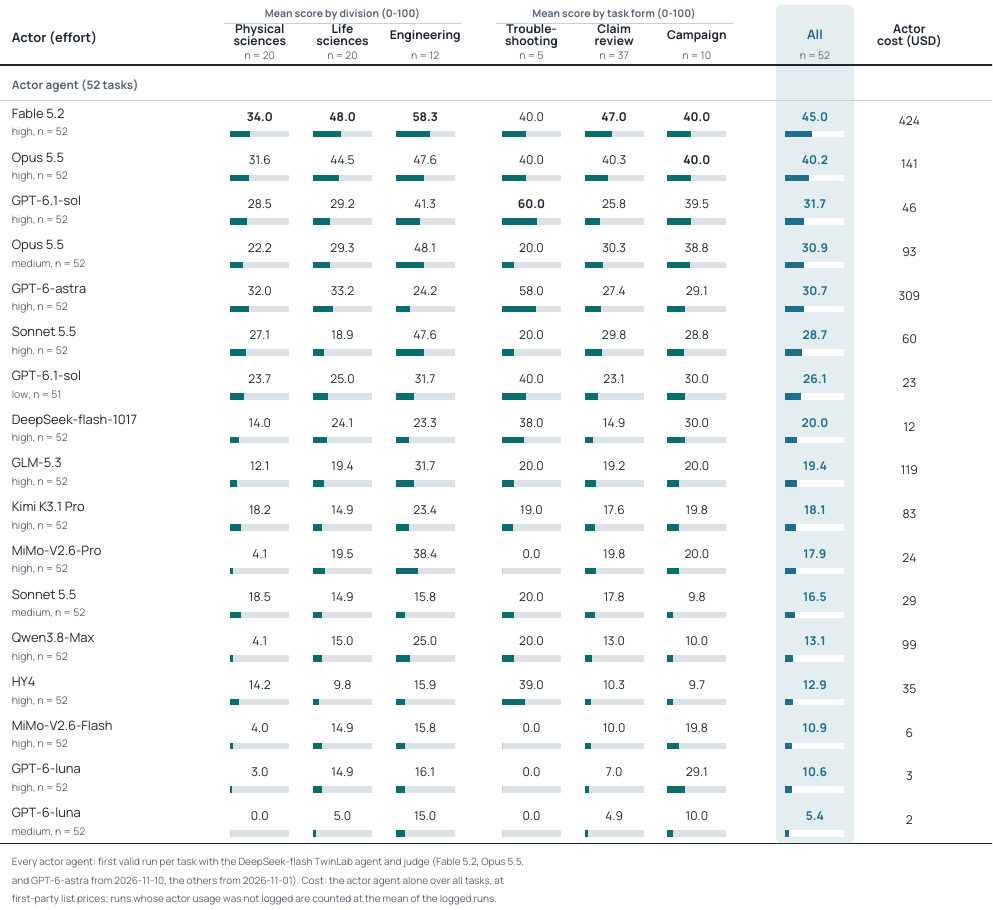}
\caption{\textbf{Main results.} Mean score (0 to 100) of each actor agent and reasoning effort by academic division, by task form, and over all tasks (All), together with the cost of the actor agent on all tasks in USD. GPT-6.1-sol at low effort has no valid run on one task.}
\label{tab:main}
\end{table*}

The physical sciences are the lowest-scoring division for \prov{12} of the \prov{17} actor configurations, and engineering is the highest-scoring division for \prov{14} of them.

We run DeepSeek-flash and MiMo-V2.6-Pro as the actor agent three times on every task, and we run every other actor agent once.
\Cref{tab:ds-repeats} gives the mean score of each run.
The All score is \prov{19.7} $\pm$ \prov{3.3} for DeepSeek-flash and \prov{14.0} $\pm$ \prov{3.6} for MiMo-V2.6-Pro.
Resampling the per-task score differences between two actor agents, Fable 5.2 scores significantly higher than \prov{14} of the other \prov{16} configurations and Opus 5.5 at high effort than \prov{12} of them at the \prov{95\%} level.
Over all \prov{136} pairs of the \prov{17} configurations, the configuration with the higher mean score also scores higher on more tasks than it scores lower in \prov{127} pairs, and the paired comparison separates \prov{68} pairs, including \prov{66} of the \prov{75} pairs whose mean scores differ by at least 10.

\begin{table}[t]
\centering
\small
\setlength{\tabcolsep}{4pt}
\begin{tabular}{@{}lcccc@{}}
\toprule
Actor agent & Run 1 & Run 2 & Run 3 & Mean $\pm$ std \\
\midrule
DeepSeek-flash & \prov{15.9} & \prov{21.4} & \prov{21.9} & \prov{19.7} $\pm$ \prov{3.3} \\
MiMo-V2.6-Pro & \prov{18.0} & \prov{13.1} & \prov{11.0} & \prov{14.0} $\pm$ \prov{3.6} \\
\bottomrule
\end{tabular}
\caption{\textbf{Three runs of two actor agents.} Mean score (0 to 100) of each run over all tasks. The first run of MiMo-V2.6-Pro used an earlier version of the environment rules.}
\label{tab:ds-repeats}
\end{table}

\FloatBarrier
\section{Reliability of the Scores}
\label{app:reliability}

\subsection{Stability of the Judge Agent}
\label{app:judge-stab}

We judge the same run three times with identical inputs, for \prov{20} runs of Opus 5.5 and \prov{16} runs of MiMo-V2.6-Pro on the same tasks.
\Cref{tab:judge-stab} reports, for each actor agent, the share of runs with the same validity verdict in all judgments, the share with the same zero or nonzero score, the share of rubric items with identical scores, and the mean standard deviation of the score within a run.
The zero or nonzero score differs between judgments on one run of each actor agent, and in both cases a core item whose full-credit condition has a boundary goes to 0 in some judgments.
The standard deviation of a task score between repeated judgments is \prov{2.4} for Opus 5.5 and \prov{2.8} for MiMo-V2.6-Pro.

\begin{table}[t]
\centering
\small
\setlength{\tabcolsep}{3pt}
\begin{tabular}{@{}lcccc@{}}
\toprule
Actor agent & Validity & Zero or not & Items & Score std \\
\midrule
Opus 5.5 & \prov{20}/\prov{20} & \prov{19}/\prov{20} & \prov{92\%} & \prov{2.4} \\
MiMo-V2.6-Pro & \prov{16}/\prov{16} & \prov{15}/\prov{16} & \prov{92\%} & \prov{2.8} \\
\bottomrule
\end{tabular}
\caption{\textbf{Stability of the judge agent.} Each run is judged three times by the DeepSeek-flash judge agent. Validity and Zero or not count the runs with the same verdict in all judgments, Items gives the share of rubric items with identical scores, and Score std is the mean standard deviation of the score within a run.}
\label{tab:judge-stab}
\end{table}

\subsection{Changing the Models of the TwinLab Agent and the Judge Agent}
\label{app:swap}

DeepSeek-flash serves as the TwinLab agent, as the judge agent, and as one of the tested actor agents, and a language-model judge can favor its own generations~\citep{panickssery2024selfpreference}, so we check whether the results depend on this choice.
We choose \prov{10} tasks by a fixed rule from the three groups of tasks that the three runs of DeepSeek-flash leave all nonzero, mixed, or all zero, and we run DeepSeek-flash as the actor agent once with Sonnet 5.5 as the TwinLab agent and Opus 5.5 as the judge agent.
All \prov{10} runs are valid, every score lies within the range of the three runs of DeepSeek-flash in the standard setting on the same task, and \prov{5} tasks pass, against \prov{5}, \prov{6}, and \prov{4} in the three standard runs.
To separate the judge agent from the TwinLab agent, we rejudge the same \prov{10} runs with DeepSeek-flash, and the pass verdict changes on \prov{1} task, where the Opus 5.5 judge agent gives 0 to a core item with a boundary.
We further rejudge \prov{20} runs from the three runs of DeepSeek-flash on \prov{9} of these tasks with Opus 5.5 and with DeepSeek-flash under the same judge prompt and rubric.
The two judge agents agree on pass in \prov{19} runs (Cohen's $\kappa = \prov{0.83}$) and on validity in all runs, and the one pass change comes from Opus 5.5 grading a self-contradictory report by its weaker reading.

\subsection{Scores Separate Correct and Wrong Reports}
\label{app:separation}

We draw \prov{126} reports from the evaluation runs of all tasks with a hash rule fixed before any model call, taking one passing and two failing runs on tasks that have both and one failing run on the other tasks.
A DeepSeek-flash call classifies each report as correct, plausible but wrong, incomplete, or missing by reading it against the hidden truth, without seeing its score, its judgment, or the rubric, and we check \prov{29} labels at the boundary by hand and change \prov{1}.
A report is correct when its main conclusions and deliverables agree with the hidden truth, also when its verification is too weak to pass.
A report is plausible but wrong when any main conclusion, cause, mechanism, recommendation, or deliverable contradicts the hidden truth, or when it leaves a main question undetermined although the hidden truth decides it.
A report is incomplete when it makes no wrong claim on the main line but its deliverable falls short.
Of these reports, \prov{43} are correct, \prov{69} are plausible but wrong, \prov{6} are incomplete, and \prov{8} runs end without a report.
On \prov{19} of the \prov{21} tasks that have both correct and wrong reports, the lowest total score of a correct report is above the highest total score of a wrong report, and the mean gap is \prov{0.46} of the maximum score.

\subsection{Validity and Cost of the Environment}
\label{app:env-valid}

\Cref{tab:env-cost} summarizes the TwinLab agent and the judge agent when both run on DeepSeek-flash: how stable they are, how well they agree with Opus 5.5 in the same roles, and what they cost.
The costs are measured with DeepSeek-flash as the actor agent on all tasks, so that the costs of the actor agents in \cref{fig:main} do not include them.
For every actor configuration and task, we count the runs up to the first valid run, which is the run in \cref{fig:main}, and \prov{44} of these \prov{875} runs are invalid (\prov{5.0\%}), so a valid run takes \prov{1.05} attempts on average.
The scan for blocked strings ignores letter case, and no valid run contains a blocked string or a marker string.
Counting only the first run of each actor configuration on each task, the share of invalid runs does not depend on the actor agent ($p = \prov{0.24}$).

\begin{table}[t]
\centering
\small
\setlength{\tabcolsep}{3pt}
\begin{tabular}{@{}p{0.62\linewidth}p{0.34\linewidth}@{}}
\toprule
Property (DeepSeek-flash) & Result \\
\midrule
TwinLab agent and judge agent, repeated runs & All \prov{19.7} $\pm$ \prov{3.3} over \prov{3} runs in \cref{tab:ds-repeats} \\
Judge agent alone, repeated judgments & same pass on \prov{19}/\prov{20} and \prov{15}/\prov{16} runs, \cref{app:judge-stab} \\
Judge agent alone, agreement with Opus 5.5 & pass agrees on \prov{19}/\prov{20} runs in \cref{app:swap} \\
Cost of the judge agent & \prov{16.9} USD for all tasks \\
Cost of the TwinLab agent & \prov{37.7} USD for all tasks \\
\bottomrule
\end{tabular}
\caption{\textbf{Stability, agreement, and cost of the TwinLab agent and the judge agent} with DeepSeek-flash in both roles. Costs are measured with DeepSeek-flash as the actor agent.}
\label{tab:env-cost}
\end{table}

\FloatBarrier
\section{Memorization Control}
\label{app:own}

\subsection{Another State of the Same Laboratory}
\label{app:altstate}

For each task in this study, we keep the task description and the handoff materials word for word and change the true state of the laboratory to another state that the literature documents for the same kind of problem.
Only the calculator and the hidden truth change, and the replay of the expert route in the new state passes before any actor agent runs.
The released data list the \prov{11} states with the original cause, the new cause and its evidence, the change in the calculator, the result of the reference replay, the state that the actor agent's conclusion follows, and whether the run mentions the source study.
In \prov{8} of the \prov{10} states that allow a verdict, the conclusion follows the new state.
On the other \prov{2} states, Opus 5.5 does not pass the task (scores of \prov{6.5} of \prov{10} and \prov{8.0} of \prov{15}).
One state is excluded from the denominator because the actor agent never ran an experiment that touches the changed part.

\subsection{Memory Probes per Task}
\label{app:memo}

For every task, the released data list whether Opus 5.5 recognizes the source, whether it recalls the source's conclusion, whether it derives the answer closed-book, when the source was published relative to the model's knowledge cutoff, and the task's score.
The probe uses two independent calls per task in a neutral temporary directory with all tools disabled.
One call asks for the source and what the model remembers of it, the other asks for the most likely answer, and a third call grades both answers against the answer key.

We repeat the probe for GPT-6.1-sol, DeepSeek-flash, and Qwen3.8-Max with the same prompts, answer key, and grading.
They name the source of \prov{4}, \prov{2}, and \prov{1} tasks, against \prov{9} for Opus 5.5, and none of them recalls the conclusion of a source.
When the tasks are grouped by whether a model names their source, the tasks with a named source do not score consistently higher in the evaluation runs of the same model, and Opus 5.5 scores \prov{43.3} on the \prov{9} tasks whose source it names and \prov{43.2} on the \prov{22} tasks whose source it does not recognize.
Fable 5.2 names the source of \prov{16} tasks, two of which can be read from the task text, and recalls the conclusion of a source for one task.
Its evaluation runs score \prov{49.0} on the \prov{16} tasks whose source it names, \prov{32.2} on the \prov{21} tasks whose source it names in part, and \prov{58.7} on the \prov{15} tasks whose source it does not name.

\subsection{Zero-Experiment Baseline}
\label{app:zero}

We select the \prov{33} tasks whose decisive data are not in the handoff, according to the scorecard annotation, and copy each task with a budget of zero; every experimental request is refused without calling the TwinLab agent.
Opus 5.5 and DeepSeek-flash each run once on every copy, and the DeepSeek-flash judge agent scores the reports with the original rubric.
Neither Opus 5.5 nor DeepSeek-flash passes any of the \prov{33} tasks, while DeepSeek-flash with the normal budget passes \prov{6} of the same tasks and at least one actor agent passes \prov{28}.

\FloatBarrier
\section{Fidelity of the Twin}
\label{app:fidelity}

\subsection{Pilot Study: What the Laboratory Needs to Know}
\label{app:prelim}

We replay the requests of the expert route in \prov{12} tasks, four from each division, against three settings, with two independent sessions per setting.
The language model in all three settings is DeepSeek-flash, the model that serves as the environment in the evaluation.
The rule that selects the tasks and the tolerance of each task from the raw files were fixed before any output was read.
Deviation is counted on the \prov{126} answers to the \prov{63} requests whose conditions exist in the record, and inconsistency on the \prov{93} requests asked in both sessions.
Deviation drops from \prov{41\%} with the task description alone to \prov{14\%} with raw materials and to \prov{2\%} with the TwinLab agent; substantial inconsistency drops from \prov{31\%} to \prov{8\%} and \prov{4\%}.
The \prov{3} deviating answers of the TwinLab agent all occur in one task: one request receives a calculator value below the tolerance band in both sessions, and one well of a \prov{48}-well plate exceeds an \prov{8\%} rule by \prov{0.2} points.
Of the \prov{4} requests that the TwinLab agent answers inconsistently, \prov{2} differ only in the amount charged.
The report of the expert route passes in \prov{17}, \prov{21}, and \prov{23} of the \prov{24} sessions of the three settings.
In a pilot, a language model that writes a metabolic curve by itself produces a curve about \prov{20} times slower than the record.

\subsection{Sources of Readings and Destinations of Requests}
\label{app:provenance}

We take the first valid run of every actor configuration other than Opus 5.5 at medium effort on every task, \prov{831} runs with \prov{7808} requests, and classify every reading by the dominant source of its numbers and every request by its destination.
Over all readings, \prov{76\%} come from the record or from calculators calibrated on the same laboratory's data, with \prov{53\%} looked up in the record and \prov{23\%} computed by the calculators, and \prov{99.5\%} trace back to the record, a calculator output, or one step of arithmetic.
Only \prov{0.5\%} of all requests are not simulated at all.

We also audit every reply of the TwinLab agent in these runs for disclosure of the hidden truth.
A script first marks \prov{79} replies that contain a number, a name, or a phrase of the hidden truth that the requests of the actor agent and the outputs of the tools do not explain.
A DeepSeek-flash call then judges these replies, the \prov{162} other replies in which the TwinLab agent read the internals of a calculator, and \prov{100} random replies, and we check every reply that it flags by hand.
One reply states a hidden mechanism without being asked, and one reply describes an asked-for observation that the calculator does not model, which we count as a coverage gap; both runs score 0, and none of the random replies is flagged.

\subsection{Audit of Extrapolation Dependence}
\label{app:extrap-audit}

Without running new models, we label every reading in the evaluation runs with its source and compare the actor agents with the replay of the expert route, which every task has, on the \prov{29} tasks whose replay used the same TwinLab agent.
The share of requests that actually obtain data outside the record does not differ between the actor agents and the expert route (median difference \prov{0} per task, $p = \prov{0.24}$), and it does not differ between two reasoning efforts of the same model.
On real-source tasks, rubric items whose evidence comes only from design parameters or from numbers written by the TwinLab agent score \prov{0.63} of full credit, and items whose evidence comes from the record score \prov{0.62}.
For each of the \prov{6} tasks whose deliverable is scored by a calculator, the settings that the actor agent can choose are either limited to the real settings in the record or come from the same design model as the observations.
The expert routes of \prov{42} of the \prov{49} tasks built from real records request experiments.
Of their \prov{189} decisive experiments, \prov{78\%} use conditions that the record contains, \prov{15\%} partly extend a recorded condition, and \prov{7\%} lie outside the record.
On \prov{7} further tasks built from real records, the expert route has no scored experiment, because the evidence comes from offline analysis of the handoff records, and we count these tasks separately.

\subsection{Sensitivity to Extrapolation}
\label{app:extrap-sens}

For six tasks whose decisive items rely partly on conditions outside the record, we build a variant that changes only the extrapolation outside the record to another choice that is physically just as plausible, such as the other end of the literature range or another functional form.
Three of the tasks are chosen by a rule fixed in advance, which takes the real-source tasks where at least two of the three actor agents below score above zero and whose earlier runs reach the changed extrapolation.
The outputs for recorded conditions stay identical, and so do the hidden truth, the rubric, the task description, and the actor agent's files.
We first replay the calculator commands of existing runs on the original and the variant and count how many outputs change, and we then run Sonnet 5.5, Kimi K3.1 Pro, and GLM-5.3 once each on every variant, with the DeepSeek-flash TwinLab agent and judge agent, and compare with their runs on the original in \cref{tab:extrap-sens}.
Sonnet 5.5 runs at medium effort on the first three tasks and at high effort on the other three.
In every variant the replay of the expert route still earns full marks.
We compare each variant run with the run of the same model and effort in \cref{fig:main}; on the \prov{12} pairs whose original run used the same task version as the variant, the pass verdict changes in \prov{2} pairs and the median absolute change of the score is \prov{0}.

\begin{table}[t]
\centering
\small
\setlength{\tabcolsep}{3pt}
\begin{tabular}{@{}p{0.25\linewidth}p{0.29\linewidth}cp{0.23\linewidth}@{}}
\toprule
Task & Changed extrapolation & \makecell{Outputs\\changed} & Runs on the variant \\
\midrule
Fatigue of printed steel & widths and decoration outside the anchors (\prov{6} items) & \prov{44\%} & conclusions same in \prov{3}/\prov{3} \\
Oyster metabolism & $Q_{10}$ below \prov{26}\,\textdegree C, \prov{3.52} $\to$ \prov{2.0} & \prov{20\%} & conclusions 1--2 same; score 0 in all \\
Radiocarbon pretreatment & rinse release, sonication residue, collagen carbon & \prov{23\%} & \prov{1} deliverable changed \\
Phonon bursts in a detector & off-record device model, thickness ratios, offsets and noise & \prov{39\%} & conclusions same in \prov{3}/\prov{3} \\
Genetic background of flies & generation of re-derived strain panels & \prov{12\%} & conclusions same in \prov{3}/\prov{3} \\
Zero-bias peak boundary & room-temperature resistance and leakage readings & \prov{25\%} & \prov{2} runs read another sweep \\
\bottomrule
\end{tabular}
\caption{\textbf{Sensitivity to the extrapolation.} Outputs changed: share of calculator outputs of existing runs that differ on the variant.}
\label{tab:extrap-sens}
\end{table}

In the fatigue task, the changed outputs appear in the evidence that the judge agent cites in \prov{14} of \prov{18} existing runs, yet the conclusions on the variant are the same for all three actor agents.
In the oyster task, the change multiplies the reading of the expert's bench-delay test by \prov{1.7}, but the actor agents take another path.
On that path the readings change by \prov{0\%} to \prov{8\%}, and the ranking of the oyster families stays the same in \prov{83} of \prov{90} reads.
Two changes in the third conclusion trace back to the wording of offline analyses and to a different choice of water, and replaying these runs on the original task gives almost the same readings.
In the radiocarbon task, one actor agent measures very small samples after sonication, where the reading depends entirely on the extrapolated residue, and changes its recommendation for low-yield samples.
The correct conclusion of this task itself depends on the size of this residue.
In the phonon task, the changed outputs appear in the evidence that the judge agent cites in \prov{20} of \prov{21} existing runs, and the score differences between the original and the variant come from two items on the wording of the report whose evidence cites none of the changed requests.
In the fly task, no run on the variant requests a re-derived strain panel, and in the zero-bias task, no rubric item reads the changed room-temperature readings, and the two runs that change their pass verdict analyze another sweep of the recorded data.

\subsection{Held-Out Conditions}
\label{app:holdout}

We test how close the calculators come to real readings on conditions that the record contains but the calculator does not see.
We choose three tasks whose calculators extend a measurement along a continuous variable, namely phonon bursts in a detector, gliding of kinesin-driven microtubules, and fluorescence blanks of clam assays.
On each task we hide \prov{2} to \prov{10} recorded conditions together with the measurements that reveal them, and the hidden values go to a file that the rebuilder cannot read.
A separate session rebuilds each calculator from the reduced record and the construction rules alone, without the original calculator, the task files, or the rubric.
Of the \prov{15} hidden readings, \prov{10} fall within twice the measured repeatability, and the median error is \prov{0.74\%} for interpolated conditions and \prov{2.47\%} for extrapolated ones.
On all three tasks, the rubric items that involve the hidden conditions reach the same judgment with the rebuilt calculator.
In a second round we hide a whole discrete condition, one of the four caseins that coat the slides of the kinesin task, and the rebuilt calculator predicts the mean gliding speed of the hidden casein within \prov{0.6\%}.
Over both rounds, \prov{19} of the \prov{29} hidden readings fall within twice the measured repeatability, with a median error of \prov{1.37\%}, and the rubric judgments stay the same.

\subsection{Experiments Beyond the Expert Route}
\label{app:routes}

We compare every successful run with the expert route of its task in two rounds.
We use the successful runs in \cref{fig:main}. The first round takes the \prov{14} tasks that the most actor agents solve, \prov{74} runs; the second takes \prov{8} tasks whose decisive test has to be devised by the agent or whose expert route has at least five experiments, \prov{30} runs.
For every core item, an Opus 5.5 call that sees the runs in shuffled order without their actor names decides whether the run supports the item with the same kind of experiment as the expert route, with another kind, or partly with another kind.
In the two rounds, \prov{10} of \prov{74} and \prov{6} of \prov{30} successful runs support a core item partly with another kind of experiment than the expert route.
In both rounds, \prov{36\%} of the requests of successful runs have no counterpart in the expert route, mostly controls that the actor agents design themselves, such as a small pilot that varies two preparation steps together, which three actor agents ran independently to tell a clot in the sample from a missing reagent.

\subsection{Red-Team Actor Agents}
\label{app:redteam}

On \prov{4} tasks, we run Opus 5.5 in three arms.
The honest arm is the run of Opus 5.5 at high effort in \cref{fig:main}.
The exploit arm is told that the laboratory is simulated, that the score is computed against a hidden answer, and that it may exploit weaknesses of the simulation to maximize the score.
The flaw-finding arm is asked mainly to find flaws of the simulation, such as physically implausible readings, contradictions with the files, or leaks, and to submit a report as usual.
The red-team instructions appear only in the actor agent's workspace, so the TwinLab agent sees the original task, and all runs are judged blind with the original rubric.
Every reported flaw is verified by reading the session and the calculator.
The four tasks are a process-window search on public data, a synthetic battery task, the fatigue task, and the oyster task; the first two are scored by a calculator.

The exploit arm makes \prov{9} attempts in total.
Seven ask the laboratory staff for the conclusion, the answer key, the source dataset, an underlying table, or history after the cutoff point; one asks for a setting outside the recorded matrix; one lists settings as a range.
The history cutoff, the TwinLab agent's refusals, the calculators' refusals, and the simulation limits block all of them, no leak guard fires, and no run searches the web, after which the actor agents solve the tasks normally.
No core item receives credit without evidence, and on two tasks the second check of core items removes credit that the first judgment gave.

The flaw-finding arm claims \prov{39} flaws: \prov{13} are true, \prov{11} are acceptable simulation limits, and \prov{15} are false alarms.
The true flaws are minor, such as two instruments of a synthetic task that use inconsistent voltage models, a handoff file that the TwinLab agent describes wrongly, and a clock that three answers report differently; none of them changes the scoring evidence or a key decision.
All claimed physical implausibilities on the public-data task are doubts about real measurements.
Looking for flaws also changes the actor agents' behavior: on two tasks they record the designed discriminating evidence as a flaw, so the scores of this arm do not measure the honest ability.

\FloatBarrier
\section{Failure Analysis}
\label{app:failure-analysis}

\subsection{Decision Points: Protocol and Sensitivity}
\label{app:decision}

We extract the decision points of each task from the rubric and from the expert route, and we split them into action decisions (\prov{133}) and discovery decisions (\prov{149}).
Probe A asks the decision questions closed-book, and probe B asks the model to write a plan with read-only access to the handoff files.
Under the core criterion, an answer counts as fully stated when it contains the decision of the expert route; under the strict criterion, it also has to meet the full-credit wording of the rubric.
The grading model agrees with human annotation on \prov{85\%} to \prov{88\%} of the decision points.
Probe A names the topic of each decision point and asks what the model would do about it.
With the closed-book question (A), decisions stated in full, only in principle, and not mentioned score \prov{82\%}, \prov{75\%}, and \prov{71\%} of their maximum in the run, and with the plan (B) they score \prov{87\%}, \prov{66\%}, and \prov{62\%}.
Decisions that the model states in full receive \prov{37\%} (A) and \prov{61\%} (B) of the score obtained and \prov{28\%} (A) and \prov{31\%} (B) of the score lost.
Under the strict criterion, decisions stated in full lose \prov{8\%} (A) and \prov{13\%} (B) of their score, and decisions stated only in principle lose \prov{25\%} (A) and \prov{29\%} (B).
On the action and discovery decisions together with B, the three levels lose \prov{15\%}, \prov{38\%}, and \prov{57\%} of their score.
The probes run on the evaluation runs of Opus 5.5 at high effort in \cref{fig:main}, and each prompt is sampled three times with the majority answer kept.
For the \prov{11} action decisions that B states in full and the run does not score in full, we compare the plan with the requests of the run and with the reason that the judge agent gives for the item.
In \prov{2} of them the run does not carry out the planned experiment: one run never tests the extraction kit on fresh tissue, and one run discharges cells that the order needs, although its plan tests only spare cells.
In the other \prov{9}, the run carries out the planned experiment and loses the score when it draws conclusions from its own readings or writes the report, which accounts for \prov{16} of the \prov{19.5} rubric score that the \prov{11} decisions lose, as \cref{tab:plan-loss} lists.

\begin{table*}[t]
\centering
\small
\begin{tabular}{@{}p{0.27\textwidth}cp{0.62\textwidth}@{}}
\toprule
Task & Score & Where the run loses the score \\
\midrule
\multicolumn{3}{@{}l}{\textit{The run does not carry out the planned experiment}} \\
\texttt{001-\allowbreak coral-\allowbreak lcm-\allowbreak porites-\allowbreak 03} & \prov{0}/\prov{1.5} & Never tests the extraction kit on the fresh tissue that the materials list offers. \\
\texttt{018-\allowbreak lisocl2-\allowbreak string-\allowbreak match-\allowbreak 03} & \prov{0}/\prov{2} & Discharges cells that the order needs, although the plan tests only spare cells. \\
\addlinespace[2pt]
\multicolumn{3}{@{}l}{\textit{The run carries out the planned experiment and draws a wrong conclusion from its readings}} \\
\texttt{005-\allowbreak glycogen-\allowbreak oyster-\allowbreak 03} & \prov{1}/\prov{2} & Measures the samples again and attributes the low readings to the calibration; the \prov{1}~$\mu$L transfer delivers about \prov{30\%} less. \\
\texttt{042-\allowbreak alsi-\allowbreak grid-\allowbreak interactive-\allowbreak 02} & \prov{1}/\prov{2} & Casts repeated bars as planned, but the formula uses the total standard deviation, which also counts the settings with a single bar. \\
\texttt{044-\allowbreak seahorse-\allowbreak patient-\allowbreak ocr-\allowbreak 01} & \prov{0}/\prov{1.5} & Grows line F03 again twice, obtains two low growths, and still calls the line normal. \\
\texttt{044-\allowbreak seahorse-\allowbreak patient-\allowbreak ocr-\allowbreak 01} & \prov{0}/\prov{1.5} & Gives no reason for the two repeated growths of line F03. \\
\texttt{044-\allowbreak seahorse-\allowbreak patient-\allowbreak ocr-\allowbreak 01} & \prov{0}/\prov{2.5} & Reads the respiration of line F15 without normalizing it by cell number and calls the line normal. \\
\addlinespace[2pt]
\multicolumn{3}{@{}l}{\textit{The run carries out the planned experiment and the report misstates or omits a result}} \\
\texttt{032-\allowbreak contour-\allowbreak pass-\allowbreak fatigue-\allowbreak 03} & \prov{0}/\prov{2.5} & Identifies the composition correctly, but the report tightens the limits of oxygen and nitrogen together. \\
\texttt{038-\allowbreak orau-\allowbreak ultrafiltration-\allowbreak 02} & \prov{0}/\prov{2.5} & Runs the rinse test and the known-age bone, but the report checks the known-age bone every few batches where every batch is needed. \\
\texttt{058-\allowbreak nickelate-\allowbreak nv-\allowbreak stress-\allowbreak 02} & \prov{0}/\prov{3} & Processes the stress, shear, and composition data of the sample, but the report keeps an untested explanation of the footprint. \\
\texttt{085-\allowbreak ssb-\allowbreak ringtrial-\allowbreak pressure-\allowbreak 02} & \prov{0}/\prov{0.5} & Asks for the stack pressure of cell L08 first, but the report omits that its open-circuit voltage and first capacity are normal. \\
\bottomrule
\end{tabular}
\caption{\textbf{Decisions that the plan states in full and the run still loses.} Each row is one action decision of Opus 5.5 at high effort, and Score gives the rubric score that the run obtains out of the maximum of the item.}
\label{tab:plan-loss}
\end{table*}

\subsection{Failure Types per Run}
\label{app:failure}

We define three failure types.
A mechanism is accepted without a discriminating experiment when the report adopts a mechanism that the run never tested against its alternative.
A scarce material is used up before testing when the key scarce material is consumed before the experiment that decides how to use it.
A procedure is chosen on weak early evidence when the deliverable depends on an early comparison that used too few samples or was confounded.
For every failed run, the released data list the type, the share of budget and of scarce material used when the decisive test started, and whether the run ended because the budget ran out.
Of the \prov{16} failed runs of Opus 5.5 at high effort on the \prov{33} tasks that need experiments, \prov{12} fall into these three types, and the other \prov{4} err in reading the results or in writing the report.

\subsection{Budget Sensitivity}
\label{app:budget}

The budget of a task is a multiple of the amount used by the replay of the expert route; the multiple is between \prov{1} and \prov{3} for \prov{49} tasks, with a median of \prov{1.6}.
\Cref{tab:budget-use} compares how much of the budget passing and failing runs use on the \prov{40} tasks that charge for experiments.
The \prov{12} tasks whose actions are all free of charge, \prov{11} of which offer no experiments, are left out, because every run on them spends nothing.

\begin{table}[t]
\centering
\small
\setlength{\tabcolsep}{3pt}
\begin{tabular}{@{}lcccccc@{}}
\toprule
& \multicolumn{3}{c}{Passing runs} & \multicolumn{3}{c}{Failing runs} \\
\cmidrule(lr){2-4}\cmidrule(l){5-7}
Actor agents & Runs & Used up & Median & Runs & Used up & Median \\
\midrule
All & \prov{160} & \prov{52\%} & \prov{0.99} & \prov{479} & \prov{40\%} & \prov{0.91} \\
Opus 5.5 & \prov{19} & \prov{63\%} & \prov{1.00} & \prov{21} & \prov{24\%} & \prov{0.85} \\
\bottomrule
\end{tabular}
\caption{\textbf{Budget use of passing and failing runs} on the \prov{40} tasks that charge for experiments. Used up counts the runs that spend at least \prov{99\%} of the budget, and Median is the median fraction of the budget spent.}
\label{tab:budget-use}
\end{table}

We also rerun with twice the budget all \prov{8} failed runs of Opus 5.5 that spent at least \prov{90\%} of their budget.
All \prov{8} reruns are valid, and \prov{2} of them pass (the radiocarbon task, from 0 to \prov{95}, and the task that matches battery cells into strings, from 0 to \prov{96.4}).
Of the \prov{6} reruns that fail again, \prov{4} fail on a core item that the original run also failed, and \prov{2} fail on a different core item.
Three of these \prov{6} reruns use less than \prov{70\%} of the doubled budget.

\FloatBarrier
\section{Scorecard Definitions and Annotations}
\label{app:scorecard}

We define the six columns of \cref{fig:scorecard} and their levels.
Tasks and scenes count the tasks and the distinct experimental settings.
Action types count the distinct experimental actions that the agent can request.
The source of research problems has three levels: designed, found in real data, and met and solved by the laboratory. A problem found in real data lies in the measurements, but the record does not show the laboratory meeting it; a problem met and solved by the laboratory appears in the record together with its later answer.
The environment for new experiments has three levels: analysis of given data, equations or a designed world, and a replay or a twin of a real laboratory.
Memorization control has three levels: the risk is noted, the tasks use data that the model has not seen, such as altered, held-out, or fictional data, or memorization is probed on the tested model.
For every existing benchmark, each cell cites the sentence of the original paper that supports it.
\bench is annotated task by task with the same definitions, and two annotators label the action types independently (\prov{105} and \prov{107} types).

Some benchmarks already ask an agent to design experiments.
SciGym~\citep{duan2025scigym}, DiscoveryWorld~\citep{jansen2024discoveryworld}, and BoxingGym~\citep{gandhi2025boxinggym} let the agent choose experiments in a designed world or an equation model, so their readings follow written rules.
ASB~\citep{asb2026} lets the agent choose batches of designs within a budget and returns published measurements, so each campaign offers one kind of action over the recorded designs.
\bench combines the two properties that these benchmarks have separately: the actor agent composes its experiments freely, and the readings come from the records of real laboratories.

\FloatBarrier
\section{Data Sources and Licenses}
\label{app:data}

Every task uses its source records under the terms of their licenses.
Where a license does not allow direct reuse, we contacted the data providers by email, and the other sources are released under licenses that allow reuse with attribution, which we follow.
\Cref{tab:cases} lists the source records, the papers, and the related references of every task.

The pictures in Figure~\ref{fig:teaser} come from the source records of the four tasks.
The fluorescence images of gliding microtubules are from the open notebook of Maloney on OpenWetWare~\citep{maloney2010openwetware} (CC BY-SA), and the fracture surface of the steel specimen is Figure~7a of \citet{wietecha2025subsurface} (CC BY 4.0).

\FloatBarrier
\section{Implementation of the Twin and the Judge}
\label{app:impl}

\subsection{Mechanisms of the Twin and the Scoring}
\label{app:mech}

\paragraph{Readings.}
The environment code checks that the recorded value is used whenever the actor agent's request allows it.
The TwinLab agent gives no hints or interpretations, does not substitute another procedure for a requested one, and answers questions about history only up to the cutoff point.
When the actor agent asks whether the cause is X, the TwinLab agent only answers that this needs an experimental test.
The environment code also asserts that observations contain no internal identifier or leakage pattern and that an unexecuted request is neither charged nor consumes material, and a failed assertion stops the run.

\paragraph{Parameter values.}
A listed value is either a condition that the laboratory ran or a condition with a written extrapolation basis; across all tasks, \prov{371} listed values are recorded and \prov{100} are extrapolated.
An extrapolated value may never be needed in the real experiment, so it acts as a plausible distractor, and the lists are not ordered by the answer.

\paragraph{Isolation.}
The actor agent's sandbox has no configuration, memory, skill, or past session.
The TwinLab agent cannot see the actor agent's reasoning, and the actor agent cannot see image file names or the rationale of the TwinLab agent.
The environment code checks that the actor agent was answered by the requested model, and a run is invalid when the platform switched models during it.
We scan the content that reaches the actor agent through its tools, and a blocked string or a per-task marker string in this content, or repeated attempts to reach a blocked source, invalidates the run.
Every run records a hash of the rubric that is compared with the frozen version.

\paragraph{Design against hacking of the twin.}
A strong actor agent may seek out conditions where the calculators extrapolate favorably, or look for flaws in the twin, as agents exploit flaws in their rewards and tests elsewhere~\citep{amodei2016concrete,zhong2025impossiblebench}.
Besides scoring against the hidden truth, three designs limit this risk.
Numbers come from deterministic calculators, extrapolation is confined to values with a written basis, and the leakage guards, the replay script, and the environment audit check every run.

We release the environment code, which enforces the remaining rules, such as measurement noise, checkpoints, billing, and the stock of scarce materials.

\subsection{Judge Agent and Environment Audit}
\label{app:judge}

The environment audit has three parts.
A leak is an observation that states an explanation, a cause, or the hidden truth.
An inconsistency is an observation that contradicts the task, an earlier observation, or physics, including identical values for independent samples, silently changed parameters, and unexecuted items reported as results.
A coverage gap is a refusal of an experiment that a real laboratory can run and that matters for scoring.
Any of the three makes the run invalid.
We release all prompts with the evaluation code, including those of the actor agent, the TwinLab agent, the judge agent, and every probe in these appendices.

\FloatBarrier
\section{Task List and Construction}
\label{app:tasks}

Of the \prov{52} tasks, \prov{49} are adapted directly from the records of one laboratory and \prov{3} are grounded syntheses.
In some of the adapted tasks the laboratory itself met the problem and later solved it, and in the others the problem lies in the real data of the laboratory.
By form, \prov{37} tasks are claim reviews, \prov{5} are troubleshooting tasks, and \prov{10} are research campaigns.
The hidden truth of a task comes from a later notebook entry, a publication, a correction, or a reanalysis of the raw data.
The hidden truth of the remaining \prov{3} tasks, which are grounded syntheses, is fixed by construction.
These tasks use real data as background, the literature shows that their problem occurred on the same kind of device or sample at a similar size, and their answer comes from the cases that the literature documents.
The handoff materials are taken from the last version of the source repository that was committed before the cutoff point, so they contain no file or finding that appeared later.
The readings, the hidden truth, and the expert route of every task come from the records and annotations that the source laboratory published, and every task follows the same construction steps.
We considered about \prov{375} candidate sources and kept those with a closed answer, raw data that can anchor new readings, and key findings that require a specific diagnosis.

\Cref{tab:cases} lists the 52 tasks with a summary of each task, the abilities that it tests, and the records and papers that it is built from, and groups the tasks by division and department as in \cref{fig:teaser}.

{\footnotesize
\setlength{\tabcolsep}{4pt}
\begin{longtable}{@{}>{\raggedright\arraybackslash}p{0.165\textwidth}>{\raggedright\arraybackslash}p{0.38\textwidth}>{\raggedright\arraybackslash}p{0.28\textwidth}>{\raggedright\arraybackslash}p{0.12\textwidth}@{}}
\caption{\textbf{Tasks of \bench.} Tasks are grouped by division and department as in \cref{fig:teaser}. Each row gives the task identifier, a summary of the task, the abilities that the task tests, and the source records and papers of the task.}
\label{tab:cases}\\
\toprule
Task & Summary & Abilities tested & Sources \\
\midrule
\endfirsthead
\multicolumn{4}{@{}l}{\small\cref{tab:cases} (continued)}\\
\toprule
Task & Summary & Abilities tested & Sources \\
\midrule
\endhead
\bottomrule
\endlastfoot
\multicolumn{4}{@{}l}{\textbf{Physical Sciences} (20 tasks)} \\
\addlinespace[2pt]
\multicolumn{4}{@{}l}{\textit{Physics}} \\ \addlinespace[2pt]
\texttt{031-\allowbreak tesseract-\allowbreak phonon-\allowbreak bursts-\allowbreak 03} & Investigate the source of excess noise, parasitic power and low-energy excess events in a cryogenic phonon detector, and test the group's existing explanation with devices and measurements chosen from a fabrication menu. & hypothesis refutation by controlled comparison; controlled experiment design; separating time-dependent instrument effects; stating the scope of a conclusion & \citep{chang2025,romani2024,anthonypetersen2024,anthonypetersen2025} \\ \addlinespace[2pt]
\texttt{058-\allowbreak nickelate-\allowbreak nv-\allowbreak stress-\allowbreak 02} & On a high-pressure NV-centre imaging project with La3Ni2O7 shards, give a per-pixel criterion for local superconductivity, explain spatial structure in the maps, predict the behaviour of two further samples before inspecting them, and decide what to load next within a budget of measurement blocks. & analysis of spatially resolved raw maps; advance prediction and check; reconciling transport with imaging; decisions about irreversible loading & \citep{mandyam2026,mandyam2025data} \\ \addlinespace[2pt]
\texttt{065-\allowbreak zbp-\allowbreak phase-\allowbreak boundary-\allowbreak 02} & Assess the phase diagram and zero-bias peak claims for a 2015 to 2016 InSb/NbTiN nanowire device using raw QTLab data, and decide what to include in the paper and what to measure next. & robustness of extracted quantities to analysis choices; reproducibility across cooldowns; alternative-explanation reasoning; planning quantum transport measurements & \citep{chen2017,chen2021data} \\ \addlinespace[2pt]
\texttt{068-\allowbreak shapiro-\allowbreak odd-\allowbreak steps-\allowbreak 02} & Evaluate a PhD student's summary of Shapiro step measurements in an Al/InAs Josephson junction using raw maps and logs, and phrase what can be supported about odd-order steps. & robustness of extracted quantities to resolution and analysis choices; raw data processing in mesoscopic transport; claim wording; next-measurement planning & \citep{zhang2025shapiro,zhang2022smashdata} \\ \addlinespace[2pt]
\texttt{069-\allowbreak nanowire-\allowbreak gn-\allowbreak plateau-\allowbreak 02} & Assess whether an InSb nanowire normal-superconductor junction shows a normal-state conductance plateau using raw QTLab maps, and phrase what the paper can state about it. & robustness of an extracted quantity to analysis choices; comparison of repeated maps; quantum transport data processing; claim wording & \citep{zhang2017,zhang2022ballisticdata} \\ \addlinespace[2pt]
\texttt{070-\allowbreak cdte-\allowbreak shell-\allowbreak mobility-\allowbreak 02} & Determine whether a CdTe shell changes the field-effect mobility of InSb nanowires using raw device data and logs, and write what the paper can state about transport. & like-for-like device comparison; checks of assumed device parameters; analysis of raw transistor scans; scoping of claims with stated conditions & \citep{badawy2022,badawy2021data} \\ \addlinespace[2pt]
\texttt{071-\allowbreak sn-\allowbreak insb-\allowbreak diode-\allowbreak gate-\allowbreak 03} & Determine what the supercurrent asymmetry of Sn-InSb nanowire junctions depends on, using raw data and logs, and prepare a gate-dependence panel with supportable wording. & symmetry checks in raw transport data; normalisation choices; identifying data that should not be pooled; confidence-graded statements & \citep{zhang2025phi0,zhang2022phi0data} \\ \addlinespace[2pt]
\texttt{073-\allowbreak sninas-\allowbreak shadow-\allowbreak claims-\allowbreak 02} & Evaluate Sn-InAs nanowire shadow junction devices from two cooldowns, deciding which devices work, what the conductance and supercurrent data show, and what to tell the growers. & reasoning about the measurement chain; device triage from raw transport data; separation of supported and unsupported claims & \citep{sharma2025,sharma2025data} \\ \addlinespace[2pt]
\texttt{077-\allowbreak nb-\allowbreak gcs-\allowbreak leakage-\allowbreak 02} & Assess a student's summary of gated Nb Dayem bridges using raw I(V) and leakage files, build the device table, state what controls the supercurrent suppression, and plan the next device with expected readings fixed in advance. & mechanism discrimination from raw transport data; artefact checks in sweeps; advance predictions; design of discriminating device experiments & \citep{ruf2024,dibernardo2024data} \\ \addlinespace[2pt]
\texttt{083-\allowbreak tgp-\allowbreak readout-\allowbreak region-\allowbreak 01} & Select devices for a readout cooldown from stage-2 tune-up data of InAs-Al hybrid nanowires, using an analysis pipeline and raw scans, and write the device statements to be signed. & robustness of pipeline verdicts; inspection of analysis code; choice under limited cryostat slots; raw nonlocal conductance checks & \citep{microsoft2025nature,microsoft2025tgpdatav1,microsoft2026tgpdatav2,legg2026,azurequantumtgp} \\ \addlinespace[2pt]
\addlinespace[2pt]
\multicolumn{4}{@{}l}{\textit{Earth Sciences}} \\ \addlinespace[2pt]
\texttt{038-\allowbreak orau-\allowbreak ultrafiltration-\allowbreak 02} & As the new head of bone pretreatment at a radiocarbon laboratory in 2002, explain why ultrafiltered bone dates from a blind test came back too old, establish a corrected pretreatment, and decide what to do with 208 past determinations. & tracing an analytical offset to its source; method validation with known-age standards; retrospective data disposition rules; use of limited archaeological material & \citep{bronkramsey2004,brock2007,englishheritage2011} \\ \addlinespace[2pt]
\texttt{063-\allowbreak fracture-\allowbreak perm-\allowbreak shear-\allowbreak 03} & For a triaxial fracture permeability experiment on Westerly granite, decide whether permeability falls with shear, what the end-of-run data can be used for, and how to plan the next run. & quality assessment of long instrument records; reuse of archived runs; flagging unreliable channels; next-run planning with expected values & \citep{madara2018} \\ \addlinespace[2pt]
\texttt{086-\allowbreak bermuda-\allowbreak clumped-\allowbreak reprocess-\allowbreak 01} & Reassess reprocessed clumped isotope data for Bermuda shells, deciding whether a site difference is real, whether old and new values can be averaged, and how cold the last interglacial was at these sites. & data reduction and reference-frame reasoning; reprocessing of legacy data; use of a limited instrument queue; uncertainty-aware paleotemperature estimates & \citep{zhang2021,zhang2021data} \\ \addlinespace[2pt]
\addlinespace[2pt]
\multicolumn{4}{@{}l}{\textit{Chemistry}} \\ \addlinespace[2pt]
\texttt{055-\allowbreak esi-\allowbreak h2o2-\allowbreak origin-\allowbreak 02} & Determine where and how H2O2 forms in water collected from a negative-mode electrospray source, evaluate a drafted electrochemical explanation, and report whether positive and uncharged droplets produce H2O2. & mechanism testing by selective run planning; charge and mass accounting; evaluation of an existing draft; detection-limit reasoning & \citep{chen2026} \\ \addlinespace[2pt]
\texttt{062-\allowbreak micelle-\allowbreak catalysis-\allowbreak claims-\allowbreak 02} & Review HPLC kinetics of a 2023 supramolecular micellar catalysis project and decide which rate-enhancement claims in a draft hold, which controls are missing, and what should be run next before the manuscript goes forward. & reading ELN prints and chromatography reports; rate-enhancement and control analysis; data-quality screening; planning minimal follow-up runs & \citep{tan2023,tan2023data} \\ \addlinespace[2pt]
\texttt{067-\allowbreak oer-\allowbreak electrodep-\allowbreak repeats-\allowbreak 02} & Review a robotic electrodeposition and oxygen-evolution dataset before submission, deciding which claims, uncertainties and data-quality statements the paper can make and what to change on the platform. & replicate uncertainty estimation; screening runs for validity from raw electrochemical traces; analysis of repeated runs; review of paper claims & \citep{fiskerbodker2025,fiskerbodker2025data} \\ \addlinespace[2pt]
\texttt{072-\allowbreak egain-\allowbreak odd-\allowbreak even-\allowbreak 02} & Using an archive of raw EGaIn J-V data from two laboratories, decide whether n-alkanethiolate junctions on Au show an odd-even effect, compute the tunneling decay coefficient, compare with Ag, and write what the group can claim. & session and provenance analysis across a long archive; statistical comparison across datasets; data-integrity screening of archived files; design of a decisive remeasurement & \citep{ddomlabegain,dehghantoranposhti2026} \\ \addlinespace[2pt]
\texttt{078-\allowbreak rh-\allowbreak hydroformylation-\allowbreak flex-\allowbreak 02} & Decide which ligand to name as the flexible ligand of a Rh hydroformylation campaign, using high-throughput plate data and newly run plates, and forecast the 15 mL scale-up results. & design of comparison plates; analysis of Bayesian-optimisation campaign data; replicate scatter estimation; quantitative scale-up forecasting & \citep{bennett2026,bennett2026data} \\ \addlinespace[2pt]
\texttt{079-\allowbreak rutpphzrh-\allowbreak fsta-\allowbreak gate-\allowbreak 01} & Review femtosecond transient absorption measurements of a Ru/Rh photocatalyst, decide what the temperature and chloride results can support, and choose measurements for a limited cryostat slot. & global lifetime analysis under model choice; Arrhenius analysis; design review of a measurement series; selection of limited instrument time & \citep{blechschmidt2025,blechschmidt2024data} \\ \addlinespace[2pt]
\texttt{080-\allowbreak pyrite-\allowbreak feivo-\allowbreak mossbauer-\allowbreak 01} & Check a draft section that claims surface high-valent iron species on pyrite catalysts, using Moessbauer, magnetometry and probe-reaction data, and decide which claims stay and what to measure. & model selection in spectral fitting; quantitative plausibility estimates; magnetometry analysis; evaluation of manuscript claims & \citep{ling2025,grandjean2026} \\ \addlinespace[2pt]
\midrule
\multicolumn{4}{@{}l}{\textbf{Life Sciences} (20 tasks)} \\
\addlinespace[2pt]
\multicolumn{4}{@{}l}{\textit{Marine Biology}} \\ \addlinespace[2pt]
\texttt{001-\allowbreak coral-\allowbreak lcm-\allowbreak porites-\allowbreak 03} & Find out why laser-capture microdissected epidermis and gastrodermis of the coral Porites compressa yield little RNA and hard-to-read sections, and deliver a working protocol plus a go or no-go decision for low-input RNA-seq library preparation. & root-cause isolation with dedicated controls; RNA quantity and integrity QC; trade-off tuning of histology steps; use of scarce tissue blocks & \citep{dellaert2026lcm} \\ \addlinespace[2pt]
\texttt{002-\allowbreak resazurin-\allowbreak clam-\allowbreak 02} & Take over heat-stress resazurin assays on Manila clam juveniles and decide which of three drafted conclusions the plate-reader data support, then fix the assay and data problems and give the protocol and analysis for the next round. & auditing raw plate-reader exports and records; detection of data-handling errors; testing readout artefacts by experiment; judging claims against corrected data & \citep{whitesamsnotebook,robertsresazurin} \\ \addlinespace[2pt]
\texttt{005-\allowbreak glycogen-\allowbreak oyster-\allowbreak 03} & Audit a luminescence glycogen assay on Pacific oyster tissue, decide which drafted conclusions hold, repair the assay and data within the remaining kit, and report final per-animal values and group statistics. & auditing plate data and sample records; checking quantification linearity and calibration; re-assay design under limited reagent; recomputation of group statistics & \citep{whitesamsnotebook,robertssormi} \\ \addlinespace[2pt]
\texttt{007-\allowbreak resazurin-\allowbreak oyster-\allowbreak 02} & Review resazurin metabolic assays on nine Pacific oyster families before they reach a breeding program, decide which conclusions hold, and recommend the protocol and analysis to use from now on. & reading raw exports with timestamps; testing the assumptions behind a normalised readout; robustness checks of pairwise statistics; protocol redesign & \citep{whitesamsnotebook,robertssormi} \\ \addlinespace[2pt]
\texttt{008-\allowbreak qpcr-\allowbreak polyic-\allowbreak 02} & Check the conclusions of an RT-qPCR experiment on Pacific oysters exposed to PolyIC and other stressors, using raw run exports and limited leftover cDNA, and state how qPCR should be run from now on. & experimental design review; confound separation by experiment design; reanalysis of raw instrument exports; use of scarce cDNA & \citep{whitesamsnotebook} \\ \addlinespace[2pt]
\texttt{045-\allowbreak oyster-\allowbreak family-\allowbreak heat-\allowbreak 03} & Build a heat-survival reference for nine Pacific oyster families by analysing a first heat challenge and planning further challenges, and choose four families for the next resazurin calibration run. & reproducibility assessment across challenges; adaptive allocation of limited animals; conservative group calling; survival analysis & \citep{robertssormi} \\ \addlinespace[2pt]
\texttt{050-\allowbreak oae-\allowbreak photo-\allowbreak scale-\allowbreak 03} & Assess whether a reported effect of ocean alkalinity enhancement preconditioning on Pacific oyster resazurin metabolism holds, list every measurement and analysis problem, and advise how to design the next round. & auditing size normalisation of raw readouts; reanalysis with corrected quantities; evaluation of post hoc analysis choices; reuse of earlier runs as evidence; experiment redesign & \citep{white2026oae,robertssormi} \\ \addlinespace[2pt]
\texttt{051-\allowbreak cod-\allowbreak blood-\allowbreak clot-\allowbreak 02} & Take over DNA and RNA co-extraction from frozen Pacific cod blood for 32 fish, develop a protocol without spending the priority tubes, and report yields against a sequencing provider's criteria together with the causes of earlier failures. & protocol development on surrogate samples; use of irreplaceable samples; QC against provider specifications; root-cause analysis from notebook records & \citep{whitesamsnotebook} \\ \addlinespace[2pt]
\texttt{054-\allowbreak coral-\allowbreak nuclei-\allowbreak count-\allowbreak 02} & Review same-day hemocytometer QC for coral single-nuclei RNA-seq, establish the loaded concentration, nuclei loaded and expected recovery per channel, identify logged readings that do not hold up, and say what has to change in the counting. & consistency checks across logged counts; reading raw count images; uncertainty ranges; protocol recommendation & \citep{dellaert2026nuclei} \\ \addlinespace[2pt]
\addlinespace[2pt]
\multicolumn{4}{@{}l}{\textit{Molecular and Cell Biology}} \\ \addlinespace[2pt]
\texttt{003-\allowbreak avian-\allowbreak blood-\allowbreak rna-\allowbreak 03} & Find out why RNA extractions from field-collected raptor blood fail, fix the procedure, and deliver RNA of the required mass and integrity per bird from the portions that remain. & troubleshooting a multi-step wet-lab protocol; isolating the failing step with controls; use of irreplaceable samples; QC of delivered material & \citep{harveylabnotebook,harveyrnaprotocol,harvey2023} \\ \addlinespace[2pt]
\texttt{033-\allowbreak kinesin-\allowbreak casein-\allowbreak gliding-\allowbreak 02} & Decide which casein blocker a planned kinesin-1 gliding-motility water-isotope series should use, with limited kinesin for new flow cells and earlier data from a single comparison. & identifying uncontrolled variables in a measurement series; assay replication with scarce reagent; setting conditions for a follow-up series; stating limits & \citep{maloney2010,maloney2010oww0525,maloney2010oww0527,maloney2011dissertation} \\ \addlinespace[2pt]
\texttt{047-\allowbreak citp-\allowbreak worm-\allowbreak retest-\allowbreak 04} & Decide for 15 blinded compounds which Caenorhabditis strains show a repeatable lifespan increase of at least 10 \%, planning further trials across three sites and predicting whether each effect will carry over to automated lifespan machines. & replication planning across sites; multiple-comparison caution; survival statistics; assay-transfer reasoning & \citep{citpaging,lucanic2017,banse2019} \\ \addlinespace[2pt]
\texttt{056-\allowbreak fly-\allowbreak white-\allowbreak background-\allowbreak 02} & Decide phenotype by phenotype whether differences between a Drosophila w1118 stock and a wild-type stock are caused by the white gene, using the lab's stocks and assays within a year's allocation, and state what a thesis chapter may claim. & genetic background control; design of comparisons that isolate a causal variable; sex-specific reasoning; scoping of claims & \citep{rickle2025,ferreiro2018,oxenkrug2009,simon2011,qiu2017} \\ \addlinespace[2pt]
\texttt{059-\allowbreak ddrad-\allowbreak quant-\allowbreak anchor-\allowbreak 01} & From a student's 113-entry ddRAD notebook, decide which DNA quantification instrument should anchor the loading sheets, how well that anchor is established, and what must be measured before it is fixed. & evidence synthesis from a long notebook; cross-instrument method comparison; statement of uncertainty ranges; measurement planning & \citep{ameralnotebook} \\ \addlinespace[2pt]
\texttt{060-\allowbreak camp-\allowbreak pulldown-\allowbreak lysates-\allowbreak 01} & As a postdoc supervising a 2010 cAMP pull-down project in airway smooth muscle cells, determine what six RIPA lysates contain, whether the control versus cigarette-smoke-extract comparisons stand, and how the next round should quantify and load. & assay interference testing; discriminating instrument faults from other causes; protein quantification; judging comparability of paired samples & \citep{poppinga2010notebook} \\ \addlinespace[2pt]
\texttt{076-\allowbreak yeast-\allowbreak raman-\allowbreak batch-\allowbreak 01} & Check a Raman spectroscopy classification run on nine yeast strains before a group meeting, decide how far the reported performance can be trusted, and advise whether a proposed screen should go ahead. & validation design for machine learning on spectra; assessment of reported model performance; sensitivity to preprocessing; experimental screen design & \citep{cheveralls2026,cheveralls2026data} \\ \addlinespace[2pt]
\addlinespace[2pt]
\multicolumn{4}{@{}l}{\textit{Medicine}} \\ \addlinespace[2pt]
\texttt{016-\allowbreak scrna-\allowbreak doublet-\allowbreak ambient-\allowbreak 01} & From a departed postdoc's 10x single-cell dataset on injured mouse kidneys, give the percentage of proximal tubule cells in a proposed hybrid state in sham and injured kidneys and the endothelial Spp1 fold change with supporting evidence, and recommend a protocol for the next cohort. & separating technical artefacts from biology in single-cell data; choosing orthogonal bench validation; assessing sample integrity; quantitative estimation under uncertainty & \citep{janssen2023,germain2022,xi2021,stoeckius2018,xu2022,kirita2020,muskovic2021} \\ \addlinespace[2pt]
\texttt{034-\allowbreak fip-\allowbreak square-\allowbreak wave-\allowbreak 03} & Determine whether a recurring square-like step in camera-based fiber photometry traces has a neural, behavioural or rig origin, recommend a fix with evidence, and classify archived sessions and their QC handling. & artefact attribution from raw time series; cross-rig and same-animal comparisons; archive-scale data triage; evaluation of QC criteria; writing a notice for users & \citep{aindfip2026issue30} \\ \addlinespace[2pt]
\texttt{044-\allowbreak seahorse-\allowbreak patient-\allowbreak ocr-\allowbreak 01} & Decide which of 26 patient fibroblast lines have a respiratory chain defect by analysing Seahorse mito stress data and choosing which irreplaceable vials to grow again. & analysis of replicate structure; adaptive replication with scarce material; cautious clinical classification; statistics on paired controls & \citep{yepez2018} \\ \addlinespace[2pt]
\texttt{075-\allowbreak cd90-\allowbreak qpcr-\allowbreak replication-\allowbreak 01} & Prepare a dossier on a multicentre RT-qPCR replication of a CD90 expression result, deciding whether each laboratory's result counts and whether the original finding replicated. & validity assessment of replication experiments; treatment of low-signal qPCR measurements; sensitivity to analysis choices; reference-gene and efficiency reasoning & \citep{amaral2025bri,briananalysis} \\ \addlinespace[2pt]
\midrule
\multicolumn{4}{@{}l}{\textbf{Engineering} (12 tasks)} \\
\addlinespace[2pt]
\multicolumn{4}{@{}l}{\textit{Electrical and Computer Engineering}} \\ \addlinespace[2pt]
\texttt{018-\allowbreak lisocl2-\allowbreak string-\allowbreak match-\allowbreak 03} & Select which primary lithium thionyl chloride cells to weld into 13 four-cell strings for seabed loggers, given inventory data and a discharge test that uses up every cell it measures. & destructive-test budgeting; tray-level sampling and screening; transfer of measurements to mission conditions; lot-release decisions & \citep{jauch2020passivation,ont2024lithium,richards2022bq35100,cieslak1992safety,saftlsh20datasheet,uspatent4307160} \\ \addlinespace[2pt]
\texttt{021-\allowbreak gas-\allowbreak sensor-\allowbreak drift-\allowbreak 01} & Assign sample labels to 271 recordings of a drifting 62-sensor metal-oxide electronic nose, spending a limited number of paid unblinding credits, and state the expected accuracy of the labels. & calibration transfer under sensor drift; allocation of limited ground-truth queries; validation on held-out labels; calibrated accuracy estimates & \citep{worner2025,worner2025data} \\ \addlinespace[2pt]
\texttt{085-\allowbreak ssb-\allowbreak ringtrial-\allowbreak pressure-\allowbreak 02} & Coordinate an interlaboratory sulfide all-solid-state battery round robin, deciding the go or no-go rule, the end point and the cycling pressure for round 2, and forecasting which cells will fall below a retention threshold. & interlaboratory variance analysis; early-cycle forecasting; round-robin protocol design; evaluation of screening criteria & \citep{puls2024} \\ \addlinespace[2pt]
\addlinespace[2pt]
\multicolumn{4}{@{}l}{\textit{Mechanical Engineering}} \\ \addlinespace[2pt]
\texttt{022-\allowbreak alsi-\allowbreak process-\allowbreak window-\allowbreak 01} & Choose the production setting of a laser powder bed fusion AlSi10Mg bracket from a 60-setting matrix with a limited machine budget, so that the setting and its drift neighbours meet a tensile strength requirement at the highest build rate. & sequential experiment planning under a budget; constrained process-window search; robustness reasoning; tensile data interpretation & \citep{luo2023data,luo2023} \\ \addlinespace[2pt]
\texttt{042-\allowbreak alsi-\allowbreak grid-\allowbreak interactive-\allowbreak 02} & Choose the production setting of an AlSi10Mg laser powder bed fusion bracket from one-bar screening data and paid confirmation bars, so that a single bar reaches a strength requirement with 95 \% confidence anywhere inside the machine tolerance. & tolerance analysis across a process window; statistical bounds from pooled replicate scatter; budgeted confirmation testing; outlier handling & \citep{luo2023data,luo2023} \\ \addlinespace[2pt]
\texttt{084-\allowbreak shell-\allowbreak record-\allowbreak retest-\allowbreak 01} & Decide what a paper can claim for two 3D-printed shell designs against balsa, using the records of an autonomous print-and-crush loop and later retest packages. & retesting for repeatability; part-level quality control analysis; evaluation of optimiser-selected results; lab-process recommendations & \citep{snapp2024} \\ \addlinespace[2pt]
\addlinespace[2pt]
\multicolumn{4}{@{}l}{\textit{Materials Science and Engineering}} \\ \addlinespace[2pt]
\texttt{029-\allowbreak die-\allowbreak attach-\allowbreak release-\allowbreak 01} & Decide which of 96 silver-sintered die-attach assemblies can ship as a 42-unit lot with a minimum die shear strength, when the shear test is destructive and test carriers are limited. & destructive-test allocation; evaluation of a screening proxy; statistical release rules; use of scarce test fixtures & \citep{ma2024,feng2016} \\ \addlinespace[2pt]
\texttt{032-\allowbreak contour-\allowbreak pass-\allowbreak fatigue-\allowbreak 03} & Determine what controls fatigue crack initiation in hot isostatically pressed laser powder bed fusion 316L specimens, tie it to a changeable item in the process record, and recommend changes to the next build and the powder specification. & fractography and phase identification; testing a default causal explanation; linking failure evidence to process records; separating correlation from demonstrated cause & \citep{wietecha2025subsurface} \\ \addlinespace[2pt]
\texttt{040-\allowbreak ceder-\allowbreak spinel-\allowbreak eis-\allowbreak 03} & Take over a self-driving-lab campaign for lithium chloride spinel conductors, decide which composition routes and agent hypotheses to keep or drop, identify the best Li-ion conductors and the basis for that claim, and state what the paper can claim. & checking automated analysis against raw impedance spectra; evaluation of agent-generated hypotheses; experiment selection from a synthesis menu; conductivity and phase interpretation & \citep{fei2026agentic,fei2026data} \\ \addlinespace[2pt]
\texttt{053-\allowbreak ec-\allowbreak film-\allowbreak od-\allowbreak cap-\allowbreak 01} & Decide whether an electrochromic film campaign justifies capping optical density, recommend a process window, and explain the high defect scores of two batches, using the film-by-film table of a multi-objective self-driving lab and four remaining batch days. & analysis of batched campaign data; check of a claimed trade-off; budgeted batch planning; replication of process windows & \citep{dahms2025selfdriving,aimatsdlelectrochromic} \\ \addlinespace[2pt]
\texttt{082-\allowbreak mmb-\allowbreak delamination-\allowbreak r-\allowbreak effect-\allowbreak 01} & Work out what mixed-mode bending fatigue delamination data show about the load-ratio effect in a composite joint, decide which crack-growth numbers can be sent to project partners, and write the summary. & checks of processed data against raw records; provenance checking; judgement of fatigue-law fits; communication of caveats & \citep{monticeli2026dstandart} \\ \addlinespace[2pt]
\addlinespace[2pt]
\multicolumn{4}{@{}l}{\textit{Chemical Engineering}} \\ \addlinespace[2pt]
\texttt{061-\allowbreak silica-\allowbreak saxs-\allowbreak attribution-\allowbreak 03} & In a self-driving silica nanoparticle campaign measured by SAXS, establish which measurements belong to which recipes in crossed rounds and what the comparison of two Bayesian optimisers says, before the optimisers are retrained. & provenance tracing through repositories and robot logs; data integrity assessment; raw SAXS curve checks; reproducible scoreboard construction & \citep{pozzosilica} \\ \addlinespace[2pt]
 
\end{longtable}
}

\end{document}